\documentclass[twocolumn]{autart}    

\usepackage[T1]{fontenc}
\usepackage{graphicx} 
\usepackage{epsfig} 
\usepackage{import} 
\usepackage{siunitx}
\usepackage{amsmath, amssymb, amsfonts}
\usepackage{algorithm}
\usepackage{array}
\usepackage{algpseudocode}

\usepackage{multirow}

\usepackage[pdfpagelabels,  
bookmarks,       
pdftex
]{hyperref}
\hypersetup{
	pdfborder   = 0 0 0,
	plainpages  = false,
	bookmarksnumbered = true,
}
\usepackage[nolist]{acronym}
\usepackage{balance}
\usepackage{graphics} 
\usepackage{color}
\usepackage{graphicx,import}          
\usepackage{todonotes}
\newtheorem{rmk}{Remark}

\newtheorem{lemma}{Lemma}

\begin{acronym}
	\acro{rnn}[RNN]{Recurrent Neural Network}
	\acroplural{rnn}[RNN]{Recurrent Neural Networks}
	\acro{narx}[NARX-NN]{Nonlinear Autoregressive Exogenous Neural Network}
	\acroplural{narx}[NARX-NN]{Nonlinear Autoregressive Exogenous Neural Networks}
	\acro{gru}[GRU]{{Gated Recurrent Unit}}
	\acroplural{gru}[GRU]{{Gated Recurrent Units}}
	\acro{lstm}[LSTM]{Long Short-Term Memory}
	\acro{ann}[NN]{Neural Network}	
	\acroplural{ann}[NN]{Neural Networks}
	\acro{ffnn}[FFNN]{Feedforward Neural Network}
	\acroplural{ffnn}[FFNN]{Feedforward Neural Networks}
	\acro{pinn}[PINN]{Physics-Informed Neural Network}
	\acroplural{pinn}[PINN]{Physics-Informed Neural Networks}
	\acro{gp}[GP]{Gaussian Process}
	\acroplural{gp}[GP]{Gaussian Processes}
	\acro{knn}[$K$NN]{$K$-Nearest-Neighbors}
	\acro{ilc}[ILC]{Iterative Learning Control}
	\acro{rc}[RC]{Repetitive Control}
	\acro{rl}[RL]{{Reinforcement Learning}}
	\acro{daoc}[DAOC]{{Direct Adaptive Optimal Control}}
	\acro{ml}[ML]{maschinelles Lernen}
	\acro{lwpr}[LWPR]{{Locally Weighted Projection Regression}}
	\acro{svm}[SVM]{{Support Vector Machine}}
	\acro{mcmc}[MCMC]{{Markov Chain Monte Carlo}}
	\acro{ad}[AD]{{Automatic Differentiation}}
	\acro{gmm}[GMM]{{Gaussian Mixture Models}}
	\acro{rkhs}[RKHS]{{Reproducing Kernel Hilbert Spaces}}
	\acro{rbf}[RBF]{{Radial Basis Function}}
	\acro{rbfnn}[RBF-NN]{{Radial Basis Function Neural Network}}
	\acroplural{rbfnn}[RBF-NN]{{Radial Basis Function Neural Networks}}
	\acro{node}[{Neural} ODE]{{Neural Ordinary Differential Equation}}
	\acroplural{node}[{Neural} ODEs]{{Neural Ordinary Differential Equations}}
	\acro{pdf}[PDF]{Probability Density Function}
	\acro{pca}[PCA]{Principal Component Analysis}

	\acro{kf}[KF]{Kalman Filter}	
	\acro{ekf}[EKF]{Extended Kalman Filter}
	\acro{nekf}[NEKF]{Neural Extended Kalman Filter}
	\acro{ukf}[UKF]{Unscented Kalman Filter}
	\acro{pf}[PF]{Particle Filter}
	\acro{mhe}[MHE]{{Moving Horizon Estimation}}
	\acro{rpe}[RPE]{{Recursive Predictive Error}}
	\acro{rls}[RLS]{{Recursive Least Squares}}
	\acro{slam}[SLAM]{{Simultaneous Location and Mapping}}
	
	\acro{mftm}[MFTM]{Magic Formula Tire Model}
	\acro{mbs}[MBS]{Multi-Body Simulation}
	\acro{lti}[LTI]{Linear Time-Invariant}
	\acro{cog}[COG]{Center Of Gravity}
	\acro{ltv}[LTV]{Linear Time-Variant}
	\acro{siso}[SISO]{Single Input Single Output}
	\acro{mimo}[MIMO]{Multiple Input Multiple Output}
	\acro{psd}[PSD]{Power Spectral Density}
	\acroplural{psd}[PSD]{Power Spectral Densities}
	\acro{cf}[CF]{Coordinate Frame}
	\acroplural{cf}[CF]{Coordinate Frames}

	\acro{pso}[PSO]{Particle Swarm Optimization}
	\acro{sqp}[SQP]{Sequentielle Quadratische Programmierung}
	\acro{svd}[SVD]{Singular Value Decomposition}
	\acro{ode}[ODE]{Ordinary Differential Equation}
	\acroplural{ode}[ODE]{Ordinary Differential Equations}
	\acro{pde}[PDE]{Partial Differential Equation}
	
	\acro{nmse}[NMSE]{Normalized Mean Squared Error}
	\acro{mse}[MSE]{Mean Squared Error}
	\acro{rmse}[RMSE]{Root Mean Squared Error}
	\acro{wrmse}[wRMSE]{weighted \ac{rmse}}
	
	\acro{mpc}[MPC]{{Model Predictive Control}}
	\acro{nmpc}[NMPC]{Nonlinear Model Predictive Control}
	\acro{lmpc}[LMPC]{Learning Model Predictive Control}
	\acro{ltvmpc}[LTV-MPC]{\acl{ltv} Model Predictive Control}
	\acro{ndi}[NDI]{Nonlinear Dynamic Inversion}
	\acro{ac}[AC]{Adhesion Control}
	\acro{esc}[ESC]{Electronic Stability Control}
	\acro{ass}[ASS]{Active Suspension System}
	\acro{trc}[TRC]{Traction Control}
	\acro{abs}[ABS]{Anti-Lock Brake System}
	\acro{gcc}[GCC]{Global Chassis Control}
	\acro{ebs}[EBS]{Electronic Braking System}
	\acro{adas}[ADAS]{Advanced Driver Assistance Systems}
	\acro{cm}[CM]{{Condition Monitoring}}
	\acro{hil}[HiL]{{Hardware-in-the-Loop}}
	\acro{siso}[SISO]{Single Input Single Output}
	\acro{mimo}[MIMO]{Multiple Input Multiple Output}

	\acro{irw}[IRW]{Independently Rotating Wheels}
	\acro{dirw}[DIRW]{Driven \acl{irw}}
	\acro{imes}[imes]{{Institute of Mechatronic Systems}}
	\acro{db}[DB]{Deutsche Bahn}
	\acro{ice}[ICE]{Intercity-Express}
	
	\acro{fmi}[FMI]{{Functional Mock-up Interface}}
	\acro{fmu}[FMU]{{Functional Mock-up Unit}}
	\acro{doi}[DOI]{{Digital Object Identifier}}

	\acro{ra}[RA]{Research Area}
	\acroplural{ra}[RA]{Research Areas}
	\acro{wp}[WP]{Work Package}
	\acroplural{wp}[WP]{Work Packages}
	
	\acro{fb}[FB]{Forschungsbereich}
	\acroplural{fb}[FB]{Forschungsbereiche}
	\acro{ap}[AP]{Arbeitspaket}
	\acroplural{ap}[AP]{Arbeitspakete}
	\acro{abb}[Abb.]{Abbildung}
	\acro{luis}[LUIS]{Leibniz Universit"at IT Services}
	
	\acro{res}[RES]{Renewable Energy Sources}
	\acro{pkw}[PKW]{Personenkraftwagen}
	\acro{twipr}[TWIPR]{{Two-Wheeled Inverted Pendulum Robot}}
	
	\acro{pmcmc}[PMCMC]{Particle Markov Chain Monte Carlo}
	\acro{mcmc}[MCMC]{Markov Chain Monte Carlo}
	\acro{rbpf}[RBPF]{Rao-Blackwellized Particle Filter}
	\acro{pf}[PF]{Particle Filter}
	\acro{ps}[PS]{Particle Smoother}
	\acro{smc}[SMC]{Sequential Monte Carlo}
	\acro{csmc}[cSMC]{conditional SMC}
	\acro{mh}[MH]{Metropolis Hastings}
	\acro{em}[EM]{Expectation Maximization}
	\acro{slam}[SLAM]{Simultaneous Location and Mapping}
	\acro{dof}[DOF]{Degree of Freedom}
	\acroplural{dof}[DOF]{Degrees of Freedom}
	\acro{pg}[PG]{Particle Gibbs}
	\acro{pgas}[PGAS]{Particle Gibbs with Ancestor Sampling}
	\acro{mpgas}[mPGAS]{marginalized Particle Gibbs with Ancestor Sampling}
	\acro{hmm}[HMM]{Hidden Markov Model}
	
	\acro{emps}[EMPS]{Electro-Mechanical Positioning System}

\end{acronym}

\newcommand{\eg}{e.\,g.,\,}
\newcommand{\ie}{i.\,e.,\,}
\newcommand*{\tr}{^{\top}}
\newcommand*{\R}{\mathbb{R}}

\newcommand*{\IW}{\mathcal{IW}}
\newcommand*{\T}{\mathcal{T}}
\newcommand*{\N}{\mathcal{N}}

\newcommand{\bi}[1]{\boldsymbol{#1}}

\newcommand\norm[1]{\left\lVert#1\right\rVert}

\DeclareMathOperator{\Tr}{Tr}

\begin{document}
	
	\begin{frontmatter}

		\title{Bayesian Inference and Learning in Nonlinear Dynamical Systems: A Framework for Incorporating Explicit and Implicit Prior Knowledge\thanksref{footnoteinfo}} 
		
		\thanks[footnoteinfo]{The material in this paper was not presented at any conference.}
		\thanks[firstauthor]{The authors contributed equally to this work and the order of authorship has been decided by tossing a coin.}
		
		\author[imes]{{Bj{\"o}rn Volkmann}\thanksref{firstauthor}}\ead{\href{mailto:volkmann@imes.uni-hannover.de}{volkmann@imes.uni-hannover.de}}, %
		\author[imes]{Jan-Hendrik Ewering\thanksref{firstauthor}}\ead{\href{mailto:ewering@imes.uni-hannover.de}{ewering@imes.uni-hannover.de}}, %
		\author[imes]{Michael Meindl}, 
		\author[imes]{Simon F.\,G. Ehlers}, 
		\author[imes]{Thomas Seel} 
		
		\address[imes]{Institute of Mechatronic Systems, Leibniz Universit{\"a}t Hannover, 30823 Garbsen, Germany}


		\begin{abstract} 
			Accuracy and generalization capabilities are key objectives when learning dynamical system models. To obtain such models from limited data, current works exploit prior knowledge and assumptions about the system. However, the fusion of diverse prior knowledge, \eg partially known system equations and smoothness assumptions about unknown model parts, with information contained in the data remains a challenging problem, especially in input-output settings with latent system state. In particular, learning functions that are nested \textit{inside} known system equations can be a laborious and error-prone expert task.
			This paper considers inference of latent states and learning of unknown model parts for fusion of data information with different sources of prior knowledge.
			The main contribution is a general-purpose system identification tool that, for the first time, provides a consistent solution for both, online and offline Bayesian inference and learning while allowing to incorporate explicit and implicit prior system knowledge. We propose a novel interface for combining known dynamics functions with a learning-based approximation of unknown system parts. 
			Based on the proposed model structure, closed-form densities for efficient parameter marginalization are derived. 
			No user-tailored coordinate transformations or model inversions are needed, making the presented framework a general-purpose tool for inference and learning. 
			The broad applicability of the devised framework is illustrated in three distinct case studies, including an experimental data set.
		\end{abstract}
		
	\end{frontmatter}

	\section{Introduction}
	An accurate system model as well as information about a system's latent states are often crucial for the design and operation of intelligent, autonomous, and adaptive real-world systems. 
	Examples of such settings include safe autonomous driving with unknown road friction conditions \cite{Volkmann.2023}, navigation in unknown environments \cite{Kok.2024}, and control systems with highly complex dynamics, \eg soft robots \cite{Mehl.2024}. 
	Two key problems that these real-world applications have in common are that measurement data is usually rare and only available in input-output settings, leaving latent (state) quantities unknown, and that 
	the underlying system dynamics are often changing during operation and/or too complex to derive from first principles, 
	resulting in inaccurate and poorly generalizing models for control and estimation.\\
	%
	%
	To tackle these challenges, current research attempts to fuse learning-based system representations with prior knowledge, \eg smoothness assumptions or known system equations. However, while quite some methods are available to consider a certain type of prior knowledge, only few existing works allow to flexibly incorporate diverse prior information.\\
	For a short review of the state of research, we consider only works concerned with input-output measurement data and introduce two main categorizations. First, we distinguish between \textbf{explicit} prior knowledge, \eg known physical equations \cite{Wigren.2022,Schon.2011}, and \textbf{implicit} prior knowledge, \eg smoothness assumptions and/or symmetry properties of the target function \cite{Geist.2021,Lutter.2023,Rath.2022}. Second, depending on the availability of measurement data, two common inference settings are introduced: \textbf{Offline system identification} using previously measured data \cite{Ljung.2010,Wigren.2022,Schon.2011}, and \textbf{online parameter estimation} using current measurement data \cite{Sarkka.2023,Ozkan.2013}.\\
	%
	In \textbf{offline system identification}, the traditional setting is to define a parametric system representation from explicit prior model knowledge, \eg by derivation from first principles. As parameter inference in nonlinear state-space models typically requires solving for a (latent) state trajectory, corresponding identification schemes often build on \ac{smc} \cite{Wills.2023} to cope with the high-dimensional search space \cite{Wigren.2022,Schon.2011}. To incorporate implicit knowledge about the parameters to be inferred, a suitable parameter prior or a regularization term can be defined \cite{Wigren.2022}. However, recent system identification schemes usually rely on a predefined parametric system model, and incorporating more flexible learning-based representations remains a challenge. As stated by \cite{Wigren.2022}, ``a key question is how to strike the right balance between prior knowledge and new knowledge via the data.''\\
	If little explicit model knowledge is available, learning-based representations can be employed to approximate the system dynamics. 
	Particularly interesting contributions for data-efficient learning build on \ac{gp} \cite{Rasmussen.2005} state-space models \cite{Frigola.2015,Frigola.2013,Svensson.2017}. 
	In \cite{Svensson.2017,Svensson.2016}, a nonlinear state-space model based on a reduced-rank \ac{gp} approximation \cite{Solin.2020} is learned. Doing so, the choice of the kernel function and its parameters allows the user to incorporate implicit knowledge about the function to be learned, \eg smoothness properties, leading to a data-efficient system identification tool for offline settings that relies on \ac{pmcmc}.
	However, while the approaches have shown promising results for simultaneous learning of state-space models and inference of latent state trajectories, interfacing other types of prior knowledge within the system identification tools, \eg explicitly known model equations, remains an expert task.\\
	%
	%
	In \textbf{online state and parameter estimation}, using explicit model knowledge is usually inevitable as inference of too many unknown quantities is not feasible with limited and sequentially available measurements. Yet, the system model for estimation can be obtained from offline learning \cite{Wolff.2024}. An interesting contribution without this restriction is presented in \cite{Berntorp.2021}, extending the results on \ac{gp} state-space modeling of \cite{Svensson.2017,Svensson.2016} to online inference and learning. In particular, the employed reduced-rank \ac{gp} \cite{Solin.2020} relies on a parametric approximation using harmonic basis functions, which is exploited in \cite{Solin.2020} to formulate recursive parameter update equations. To enable learning of the target function and inference of latent states, a marginalized \ac{pf} is tailored. However, large numbers of required basis functions and associated weighting parameters can easily raise observability/identifyability problems, considering the limited measurement data available for online learning. This calls for incorporating further prior knowledge to guide the learning process. For instance, it is proposed in \cite{Berntorp.2022} to introduce linear operator constraints or to employ user-defined basis functions for encoding, \eg symmetry knowledge about the function to be learned. If such expert knowledge is not available, a set of expressive basis functions can be conditioned offline in a data-driven fashion to enable efficient online learning with few parameters to determine \cite{Ewering.2024b}. However, although embedding explicitly known system equations in \cite{Berntorp.2021} is possible, constructing such a ``hybrid'' model and tailoring the corresponding \ac{smc} scheme can be challenging.\\
	On the other hand, implicit knowledge about a system's inherent structure can be exploited for marginalization of states or parameters, facilitating accuracy and efficiency in inference schemes. In \cite{Schon.2005}, for instance, linear substructures in the system dynamics are exploited for marginalization in \ac{pf}s. Similarly, \cite{Ozkan.2013} place suitable priors on noise parameters and leverage conjugacy properties to obtain an adaptive marginalized \ac{pf}. While these approaches provide vital schemes for efficient inference, finding or tailoring the required model structure in a particular application can be difficult and considers only one specific type of prior knowledge.\\ 
	%
	%
	In summary, existing works provide useful methods for individual classes of nonlinear system identification tasks with specific types of prior knowledge. However, it remains a major challenge to fuse multiple, diverse types of prior knowledge -- \eg explicitly known equations and implicit expert assumptions about a function to be learned -- in a joint modeling and identification framework. 
	Moreover, learning functions that are nested \textit{inside} known system equations frequently rely on user-defined coordinate transformations or model inversions, which makes derivation and identification of such hybrid models a laborious and error-prone expert task.
	Finally, existing methods usually provide only partial solutions for a specific inference setting, leaving the question of a consistent identification tool for both, online and offline applications open.\\
	%
	In this paper, we aim to overcome these three limitations by proposing a general-purpose system identification tool that, for the first time, provides a consistent solution for both, online and \textbf{online and offline inference and learning} while allowing to incorporate explicit and implicit prior system knowledge. The framework relies on two components: (1) a novel model structure for simple and interchangeable interfacing of a flexible basis function expansion with explicit and implicit prior system knowledge without the need for user-defined coordinate transformations or model inversions; (2) tailored \ac{smc} methods that exploit the model structure for efficient inference of latent states and learning of the target function \textit{inside} a partially known nonlinear state-space model.
	%
	%
	For inference and learning in the proposed model structure, closed-form expressions of the parameter densities are derived and exploited in a marginalized \ac{pf} and a marginalized \ac{pmcmc} scheme, thus extending existing results \cite{Svensson.2017,Berntorp.2021}. The resulting \ac{pf} comes with the usual asymptotic performance guarantees and the derived \ac{pmcmc} algorithm by construction asymptotically provides samples from the true parameter posterior \cite{Andrieu.2010,Svensson.2017}. The proposed method is validated in three different case studies, including an experimental benchmark of an Electro-Mechanical Positioning System (EMPS).
	%
		
		\textbf{Notation:} For a vector $\bi{e} \in \R^{n_e}$, $\bi{e} \sim \N\left(\bi{\mu},\bi{\Sigma}\right)$ denotes a draw from a multivariate Normal with mean $\bi{\mu}$ and covariance $\bi{\Sigma}$, and $e_i$ is its $i$-th element. $\norm{\bi{e}}$ denotes the L$2$-norm of $\bi{e}$. We use column vectors if not stated explicitly otherwise. 
		A matrix $\bi{A}$ is written in bold and capital and has elements $a_{ij}$ for row $i$ and column $j$. 
		The identity matrix of dimension $n$ is $\mathbf{I}_n$. 
		We denote $|\bi{A}|$ as the determinant and $\Tr(\bi{A})$ as the trace of matrix $\bi{A}$.
		We write the conditional density of a state sequence $\bi{x}_{1:t} := \{ \bi{x}_i \}_{i=1}^t$ from time steps $1$ to $t$, given the measurements $\bi{y}_{1:t}$ as $p(\bi{x}_{1:t}|\bi{y}_{1:t})$. 
		The notation $\bi{\theta}_{t \mid t-1}$ refers to the one-step prediction of variable $\bi{\theta}$ from time step $t-1$, and $\bi{\theta}_{t \mid t}$ is the posterior variable, after incorporating the evidence of time step $t$. 
		The matrix normal distribution $\mathcal{MN}(\bi{M}, \bi{\Sigma}, \bi{V})$ has mean matrix $\bi{M}$, column-covariance matrix $\bi{V}$, and row-covariance matrix $\bi{\Sigma}$. 
		The inverse Wishart distribution with scale matrix $\bi{\Psi}$, and degree of freedom $\nu$ is $\IW(\nu,\bi{\Psi})$. 
		The multivariate Student-t distribution is $\T\left(\nu, \bi{\mu},\bi{\Psi}\right)$. 
		The Dirac delta mass $\delta_{i}(j) = 1$ for $i=j$ and $0$ otherwise. 
		
		\textbf{Outline:} The paper is structured as follows. First, the problem is formalized in Section~\ref{ch:problem}. 
		Section~\ref{ch:MdlStr} introduces the overall learning framework, including the proposed model structure and the general inference architecture. In Section~\ref{ch:methods_online} and \ref{ch:methods_offline}, the offline and online inference and learning methods are derived and explained in detail, respectively. Section~\ref{ch:results} presents numerical and experimental results, including inference and learning in a real-world battery system. Finally, Section~\ref{ch:conclusion} concludes the paper.

		\section{Problem statement}\label{ch:problem}
		%
		%
		%
		%
		%
		%
		%
		Consider a nonlinear dynamical system with latent variables and system outputs, \eg an autonomous car, or an industrial robot. Building on first principles, nominal models of such systems are often available. However, complex real-world systems usually exhibit further dynamics that are hard to derive from first principles, such as friction characteristics between two surfaces, or the stiffness of a soft actuated robot. Though, some intuition about the considered effects, \eg smoothness assumptions, symmetry, or boundary conditions, may be available from expert knowledge.\\ 
		%
		%
		Formally, we consider a nonlinear stochastic discrete-time state-space system of the form
		\begin{subequations}\label{eq:problem}
			\begin{align}
				\bi{x}_{t+1} &= \bi{f}(\bi{x}_{t}, \bi{\Xi}(\bi{x}_{t})) + \bi{\omega}_t, \\
				\bi{y}_{t} &= \bi{h}(\bi{x}_{t}) + \bi{e}_t,
			\end{align}
		\end{subequations}
		where the nonlinear function $\bi{\Xi}: \mathbb{R}^{n_{x}} \mapsto \mathbb{R}^{n_{\xi}}$ is unknown and describes the considered effects beyond the first principles model. The latent states $\bi{x}_t \in \Omega_{x} \subset \mathbb{R}^{n_{x}}$ are observed through the outputs $\bi{y}_t \in \mathbb{R}^{n_{y}}$ at each time step $t$. The nonlinear state dynamics $\bi{f}: \mathbb{R}^{n_{x}} \times \mathbb{R}^{n_{\xi}} \mapsto \mathbb{R}^{n_{x}}$ and the measurement functions $\bi{h}: \mathbb{R}^{n_{x}} \mapsto \mathbb{R}^{n_{y}}$ are explicit prior model knowledge. 
		The zero-mean Gaussian noise terms $\bi{\omega}_t$ and $\bi{e}_t$ are drawn $\bi{\omega}_t \sim \mathcal{N}(\bi{0},\bi{\Sigma}_{\omega})$ and $\bi{e}_t \sim \mathcal{N}(\bi{0},\bi{\Sigma}_{e})$ with known covariance matrices, respectively.\\
		The problem considered in this work is to develop a framework for learning the function $\bi{\Xi}$ \textit{inside} the nonlinear state dynamics from input-output data while guiding the learning process with implicit prior knowledge about the target function, \eg smoothness assumptions or symmetry properties. The inference task is to simultaneously estimate the latent states $\bi{x}_{t}$, as the state variables are not fully measured. 
		For offline inference and learning, a full-length measurement trajectory $\bi{y}_{0:T}$ is available. For online application, a recursive algorithm that exploits the measurements $\bi{y}_{0:t}$ up to the current time instant $t$ is to be developed. 
		\begin{rmk}
			To retain a concise notation, possible system inputs $\bi{u}_t \in \mathbb{R}^{n_{u}}$ are not considered explicitly, but extension to such settings is straightforward, as illustrated in Section \ref{ch:results} based on several case studies.
		\end{rmk}
		\section{Overview of the proposed learning framework} \label{ch:MdlStr}
		%
		The proposed learning framework comprises two major parts. First, a convenient model structure for embedding diverse prior knowledge about the considered system is introduced in Section \ref{sec:Model_structure}. Second, in Section \ref{sec:Inference_and_learning}, we outline how the model structure can be exploited for Bayesian inference and learning. The detailed algorithms for online and offline inference and learning are derived in Sections \ref{ch:methods_online} and \ref{ch:methods_offline}, respectively.
		\subsection{Model structure}\label{sec:Model_structure}
\begin{figure}
			\begin{center}  
				\fontsize{8pt}{10pt}\selectfont 
				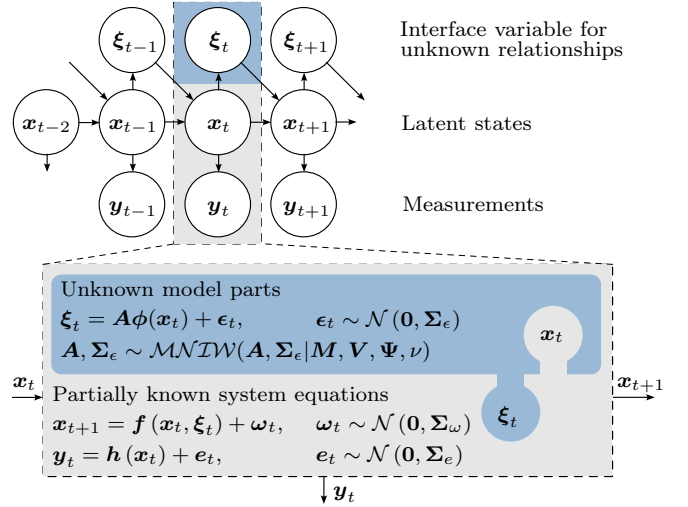
				\normalsize 
				\caption{Illustration of the identification framework that allows the incorporation of diverse prior knowledge about a system while using input-output data only. The framework relies on a novel model structure for simple and interchangeable interfacing of a flexible basis function expansion with explicit and implicit prior system knowledge.}  
				\label{fig:teaser_graphic}                                 
			\end{center}                                 
		\end{figure}

		To approach the current problem, we define an interface variable $\bi{\xi}_t \in \mathbb{R}^{n_{\xi}}$ for merging the known system equations $\bi{f}$ and $\bi{h}$ agnostically with a learning-based approximation for the unknown relationship $\bi{\Xi}$. Therefore, we parameterize the approximation $\hat{\bi{\Xi}}$ of the target function as a basis function expansion
		\begin{equation}\label{eq:basis_func_exp}
			\bi{\xi}_t = \hat{\bi{\Xi}}(\bi{x}_t) = \underbrace{\begin{bmatrix}
					a_{11} & \hdots & a_{1n_{\phi}} \\
					\vdots & & \vdots\\
					a_{n_{\xi}1} & \hdots & a_{n_{\xi}n_{\phi}}
			\end{bmatrix}}_{\bi{A}} \underbrace{\begin{bmatrix}
					\phi_1 (\bi{x}_t) \\ \vdots \\ \phi_{n_\phi} (\bi{x}_t)
			\end{bmatrix}}_{\bi{\phi} (\bi{x}_t)} + \bi{\epsilon}_t,
		\end{equation}
		with basis functions $\phi_i : \mathbb{R}^{n_x} \mapsto \mathbb{R}$, $i=1,\dots,n_{\phi}$, weight matrix $\bi{A}$, and zero-mean Gaussian noise $\bi{\epsilon}_t$ with unknown covariance $\bi{\Sigma}_\epsilon$, \ie $\bi{\epsilon}_t \sim \mathcal{N}(\bi{0}, \bi{\Sigma}_\epsilon)$.\\
		Thus, the explicit model knowledge $\bi{f}$ and $\bi{h}$ is merged with the approximation \eqref{eq:basis_func_exp} using the interface variable $\bi{\xi}_t$, and system \eqref{eq:problem} becomes
		\begin{subequations} \label{eq:MdlStr_ModelEquationsSSM}
			\begin{align}
				\bi{x}_{t+1} &= \bi{{f}} \left(\bi{x}_{t}, \bi{\xi}_{t} \right) + \bi{\omega}_t, \label{eq:state_trans}\\
				\bi{\xi}_{t} &= \bi{A} \bi{\phi}(\bi{x}_{t}) + \bi{\epsilon}_t,\\
				\bi{y}_{t} &= \bi{h}\left(\bi{x}_{t} \right) + \bi{e}_t,
			\end{align}
		\end{subequations}
		where the learning problem consists in finding the parameters $\bi{\theta}=\{ \bi{A}, \bi{\Sigma}_{\epsilon}\}$. The resulting model structure is illustrated in Fig.~\ref{fig:teaser_graphic}.\\
		To facilitate learning, formulation \eqref{eq:basis_func_exp} allows to incorporate implicit knowledge about $\bi{\Xi}$ in at least two ways.
		First, following a Bayesian approach, knowledge about $\bi{\theta}$ can be incorporated by setting a suitable (conjugate) prior on the parameters. For learning, choosing a conjugate prior has the beneficial effect that the parameter posterior is available in closed form. For the current model structure \eqref{eq:basis_func_exp}, a matrix-normal inverse Wishart ($\mathcal{MNIW}$) prior 
		\begin{equation} \label{eq:MdlStr_MNIW_Prior}
			\begin{aligned}
				\bi{A}, \bi{\Sigma}_{\epsilon} \sim & \, \mathcal{MNIW}(\bi{A}, \bi{\Sigma}_{\epsilon}| \bi{M}, \bi{V}, \bi{\Psi}, \nu) \\
				& =\mathcal{MN}(\bi{A}|\bi{M},  \bi{\Sigma}_{\epsilon}, \bi{V})\mathcal{IW}(\bi{\Sigma}_{\epsilon}| \bi{\Psi}, \nu),
			\end{aligned}
		\end{equation}
		is a suitable choice, as shown for related settings in \cite{Svensson.2017,Berntorp.2021}. The matrix-normal ($\mathcal{MN}$) distribution \cite{Dawid.1981} is
		\begin{equation}
			\begin{aligned}
				&\mathcal{MN}(\bi{A}|\bi{M},  \bi{\Sigma}_{\epsilon}, \bi{V}) = \frac{1}{(2 \pi)^{n_{\xi} n_{\phi} / 2}|\bi{V}|^{n_{\xi} / 2}|\bi{\Sigma}_{\epsilon}|^{n_{\phi} / 2}} \\
				& \; \cdot \exp \left(-\frac{1}{2} \Tr \left((\bi{A}-\bi{M})\tr \bi{\Sigma}_{\epsilon}^{-1}(\bi{A}-\bi{M}) \bi{V}^{-1}\right)\right),
			\end{aligned}
		\end{equation}
		and the inverse Wishart ($\mathcal{IW}$) distribution is
		\begin{equation}
			\begin{aligned}
				&\mathcal{IW}(\bi{\Sigma}_{\epsilon}| \bi{\Psi}, \nu) = \frac{|\bi{\Psi}|^{\nu / 2}|\bi{\Sigma}_{\epsilon}|^{-\left(n_{\xi}+\nu+1\right) / 2}}{2^{\nu n_{\xi} / 2} \bi{\Gamma}_{n_{\xi}}(\nu / 2)} \\
				& \hspace{3.5cm} \cdot \exp \left(-\frac{1}{2} \Tr \left( \bi{\Sigma}_{\epsilon}^{-1} \bi{\Psi}\right)\right),
			\end{aligned}
		\end{equation}
		with $\bi{\Gamma}_{n_{\xi}}(\cdot)$ being the multivariate gamma function.\\
		Second, the basis functions $\bi{\phi}(\bi{x}_t)$ can be chosen at custom to match the requirements of the application at hand. If implicit prior knowledge, such as assumptions regarding symmetry or smoothness, is available, learning may be intentionally biased by defining a library of feasible basis functions. 
		If no implicit prior knowledge regarding $\bi{\phi}(\bi{x}_t)$ is available, generic activation functions or harmonic features can be employed to flexibly approximate the target function using, \eg universal function approximators.
		\begin{rmk}
			For notational clarity, a single interface variable vector $\bi{\xi}_t$ is considered in the state transition function \eqref{eq:state_trans}. However, without loss of generality, the results can be extended to account for
			\begin{enumerate}
				\item multiple (interdependent) interface variables,
				\item a measurement function $\bi{h}(\bi{x}_t,\bi{\xi}_t)$, dependent on $\bi{\xi}_{t}$, and
				\item other likelihood distributions $p(\bi{\xi}_{t}| \bi{x}_{t}, \bi{\theta}_{t})$ with corresponding conjugate parameter priors $p(\bi{\theta}_{t})$.
			\end{enumerate}
		\end{rmk}
		
		\subsection{Bayesian inference and learning approach}\label{sec:Inference_and_learning}
		To enable learning and inference without inversion of the known and possibly nonlinear system equations, we exploit a data augmentation strategy and conditionally linear substructures in the representation of the unknown model parts.\\
		Based on the proposed model structure \eqref{eq:MdlStr_ModelEquationsSSM} and given the measurements $\bi{y}_{*}$, we seek to infer the latent states $\bi{x}_{*}$ and interface variables $\bi{\xi}_{*}$ while learning the parameters $\bi{\theta}_{*}$, with the subscript $\square_{*}$ referring to $\square_{0:t}$ in the online setting and $\square_{0:T}$ for offline inference and learning, respectively. In particular, we target the joint posterior density
		\begin{align} \label{eq:MdlStr_posterior}
			p(\bi{x}_{*}, \bi{\xi}_{*}, \bi{\theta}_{*}|\bi{y}_{*}) 
			&= \underbrace{p(\bi{\theta}_{*}|\bi{x}_{*}, \bi{\xi}_{*})}_{\text{posterior (ii)}} \underbrace{p(\bi{x}_{*}, \bi{\xi}_{*}|\bi{y}_{*})}_{\text{posterior (i)}},
		\end{align}
		where we approximate posterior (i) using tailored \ac{smc} methods (see Section~\ref{ch:methods_online} and \ref{ch:methods_offline}). By construction, the \ac{smc} algorithms provide access to the inferred states and interface variables $(\bi{x}_{*}, \bi{\xi}_{*} )$, which allows to compute the posterior (ii) in closed form, given the conjugate prior\footnote{Additionally, posterior (ii) simplifies to $p(\bi{\theta}_{*}|\bi{x}_{*}, \bi{\xi}_{*}, \bi{y}_{*}) = p(\bi{\theta}_{*}|\bi{x}_{*}, \bi{\xi}_{*})$, due to the current model structure \eqref{eq:MdlStr_ModelEquationsSSM} with $\bi{\xi}_t$ contributing to the state transition function only.}.\\

		%
		
		\section{Online inference and learning}\label{ch:methods_online}
		For online inference and learning, we are given the current measurement $\bi{y}_{t}$ at time step $t$ and target the joint filtering posterior 
		\begin{equation}\label{eq:online_posterior}
			p(\bi{x}_{0:t}, \bi{\xi}_{0:t}, \bi{\theta}_{t}|\bi{y}_{0:t}) = p(\bi{\theta}_{t}|\bi{x}_{0:t}, \bi{\xi}_{0:t}) p(\bi{x}_{0:t}, \bi{\xi}_{0:t}|\bi{y}_{0:t}),
		\end{equation}
		where the parameters $\bi{\theta}_{t}$ are modeled time-variant. 
		Following the lines of \cite{Berntorp.2021} and exploiting the decomposition in \eqref{eq:online_posterior}, we generate samples from the joint posterior by recursive evaluation of $p(\bi{\theta}_{t}|\bi{x}_{0:t}, \bi{\xi}_{0:t})$ and $p(\bi{x}_{0:t}, \bi{\xi}_{0:t}|\bi{y}_{0:t})$. In particular, the current parameter posterior $p(\bi{\theta}_{t-1}|\bi{x}_{0:t-1}, \bi{\xi}_{0:t-1})$ is constructed in closed form, given the trajectory estimates $\bi{x}_{0:t-1}, \bi{\xi}_{0:t-1}$ (Section \ref{ch:online_learning}). Conversely, new state and interface variable estimates $\bi{x}_t$, $\bi{\xi}_t$ are inferred using a marginalized \ac{pf}, given that marginalization of the parameters $\bi{\theta}$ can be done efficiently thanks to the closed-form parameter posterior (Section \ref{ch:online_inference}). 
		The resulting scheme is summarized in Algorithm \ref{alg:online_RBPF}.

		\subsection{Learning: The parameter posterior}\label{ch:online_learning}
		To construct the current parameter posterior from the state and interface variable trajectories, we decompose
		\begin{equation}\label{eq:online_param_posterior}
			\begin{aligned}
				&p(\bi{\theta}_{t}|\bi{x}_{0:t}, \bi{\xi}_{0:t}) \propto  \\
				&\quad p(\bi{x}_{t}, \bi{\xi}_{t} | \bi{\theta}_{t-1}, \bi{x}_{t-1}, \bi{\xi}_{t-1}) p(\bi{\theta}_{t-1}|\bi{x}_{0:t-1}, \bi{\xi}_{0:t-1}),
			\end{aligned}
		\end{equation}
		using Bayes' rule, where the likelihood is
		\begin{equation}\label{eq:online_parameter_decomposition}
			\begin{aligned}
				&p(\bi{x}_{t}, \bi{\xi}_{t}| \bi{\theta}_{t-1}, \bi{x}_{t-1}, \bi{\xi}_{t-1})  \\
				&\qquad = p( \bi{\xi}_{t} \mid \bi{x}_{t}, \bi{\theta}_{t-1} ) p(\bi{x}_{t} \mid  \bi{x}_{t-1}, \bi{\xi}_{t-1}).
			\end{aligned}
		\end{equation}
		Because we assume that the state transition function \eqref{eq:state_trans} is known and, thus, independent of $\bi{\theta}$, \eqref{eq:online_parameter_decomposition} is proportional to $p(\bi{\xi}_{t} | \bi{x}_{t}, \bi{\theta}_{t-1})$ regarding the parameters. Given the Gaussian noise model 
		\begin{equation}
			\begin{aligned}
				p( \bi{\xi}_{t} \mid \bi{x}_{t}, \bi{\theta}_{t-1}) = \mathcal{N}\left(\bi{\xi}_{t} \mid \bi{A} \bi{\phi}(\bi{x}_{t}), \bi{\Sigma}_\epsilon \right),
			\end{aligned}
		\end{equation}
		and due to the choice of a conjugate prior, the parameter posterior can be calculated recursively in closed form by summation of the previous sufficient statistics and the new statistics using the estimates $\bi{x}_{t}$, $\bi{\xi}_{t}$. This property holds due to the following Lemma\,\ref{def:tranformation_suff_stats}, relating distribution parameters and sufficient statistics, and Theorem\,\ref{thm:Online}, providing the update law that enables analytical marginalization.
		
		\begin{algorithm}[tb]
			\caption{Marginalized particle filter for online inference and learning of \eqref{eq:MdlStr_ModelEquationsSSM} (for all $i=1,\dots,N$)}
			\label{alg:online_RBPF}
			\textbf{Initialize:} Data $\bi{y}_{0:T}$, parameter prior $p(\bi{\theta})$, particles $\{ \bi{x}_{0}^{i},\bi{\xi}_{0}^{i} \}_{i=1}^{N} \sim p( \bi{x}_0,\bi{\xi}_0|\bi{\theta})$, weights $\{q^i_{0}\}_{i=1}^{N} = 1/N$.
			\begin{algorithmic}[1]
				\For{$t = 1,\dots,T$}
				\State Statistics time update
				\Statex \hspace{4mm} $\bi{\eta}^i_{t \mid t-1} \leftarrow \bi{\eta}^i_{t-1 \mid t-1}$. \Comment{by \eqref{eq:online_time_update}}
				\State Generate auxiliary states $\tilde{\bi{x}}_t^{i} = \bi{f}(\bi{x}_{t-1}^{i},\bi{\xi}_{t-1}^{i})$.
				\State Compute first-stage weights 
				\Statex \hspace{4mm} $\lambda^{i} \propto q_{t-1}^{i} \mathcal{N}(\bi{y}_t | \bi{h}(\tilde{\bi{x}}_t^{i}) ,\bi{\Sigma}_e)$ and normalize.
				\State Draw $a_t^{i} \sim \mathcal{C}(\{\lambda^{i}\}_{i=1}^{N})$.
				\State Draw $\bi{x}_t^{i} \sim \mathcal{N}(\bi{x}_t|\bi{f}(\bi{x}_{t-1}^{a_t^{i}},\bi{\xi}_{t-1}^{a_t^{i}}), \bi{\Sigma}_\omega)$.
				\State Draw $\bi{\xi}_t^{i} \sim p(\bi{\xi}_t | \bi{x}_t^{i}, \bi{\eta}^{a_t^{i}}_{t \mid t-1})$. \Comment{by \eqref{eq:online_predictive_distribution}}
				\State Statistics measurement update and
				\Statex \hspace{4mm} resampling $\bi{\eta}^i_{t \mid t} \leftarrow \bi{\eta}^{a_t^{i}}_{t \mid t-1}$. \Comment{by \eqref{eq:online_meas_update}}
				\State Compute weights ${q}_t^{i}$ and normalize. \Comment{by \eqref{eq:online_weight_update}}
				\EndFor
			\end{algorithmic}
		\end{algorithm}
		\begin{lemma}\label{def:tranformation_suff_stats}
			The parameters $\bi{\Theta} = \{\bi{M}, \bi{V}, \bi{\Psi}, \nu\}$ of a matrix-normal inverse Wishart distribution\\ $\mathcal{MNIW}(\bi{M}, \bi{V}, \bi{\Psi}, \nu)$ can be calculated from the corresponding sufficient statistics $\bi{\eta} = \{ \bi{\chi}_{0},\bi{\chi}_{1},\bi{\chi}_{2},{\chi}_{3} \}$ by
			\begin{subequations} \label{eq:MdlStr_mniw_suffstat2para}
				\begin{align}
					\bi{M} &= (\bi{\chi}_{1}^{-1} \bi{\chi}_{0})\tr, & \bi{V} &= \bi{\chi}_{1}^{-1},\\
					\bi{\Psi} &= \bi{\chi}_{2} - \bi{\chi}_{0}\tr \bi{\chi}_{1}^{-1} {\bi{\chi}_{0}\tr}, & \nu &= \chi_{3},
				\end{align}
			\end{subequations}
			and the inverse relationship is
			\begin{subequations} \label{eq:MdlStr_mniw_para2suffstat}
				\begin{align}
					\bi{\chi}_{0} &= \bi{V}^{-1}\bi{M}\tr, & \bi{\chi}_{1} &= \bi{V}^{-1},\\
					\bi{\chi}_{2} &= \bi{M} \bi{V}^{-1} \bi{M}\tr + \bi{\Psi}, & \chi_{3} &= \nu.
				\end{align}
			\end{subequations}
		\end{lemma}
		\begin{pf}
			The results can be obtained by writing the considered $\mathcal{MNIW}$ distribution in the canonical form of the restricted exponential family \cite{Wigren.2019} and comparing terms (see Appendix A1). Similar results have been presented in \cite{Svensson.2017}.
		\end{pf}
		\begin{thm}\label{thm:Online}
			Given model structure \eqref{eq:MdlStr_ModelEquationsSSM}, parameter prior $p(\bi{\theta}_{t-1} | \bi{x}_{0:t-1}, \bi{\xi}_{0:t-1})$ with distribution parameters $\bi{\Theta}_{t|t-1}=\{\bi{M}_{t|t-1}, \bi{V}_{t|t-1}, \bi{\Psi}_{t|t-1}, \nu_{t|t-1}\}$, and samples $\bi{x}_{t}$, $\bi{\xi}_{t}$, the current posterior distribution of the parameters at time $t$ is
			\begin{align}
				p(\bi{\theta}_t|\bi{x}_{0:t}, \bi{\xi}_{0:t}) = \mathcal{MNIW}(\bi{A}, \bi{\Sigma}_{\epsilon}| \bi{M}_{t \mid t}, \bi{V}_{t \mid t}, \bi{\Psi}_{t \mid t}, \nu_{t \mid t}),
			\end{align}
			with $\bi{M}_{t \mid t}$, $\bi{V}_{t \mid t}$, $\bi{\Psi}_{t \mid t}$, and $\nu_{t \mid t}$ being calculated by \eqref{eq:MdlStr_mniw_suffstat2para}, if the corresponding statistics $\bi{\eta}_{t|t}$ are recursively updated as
			\begin{subequations}\label{eq:online_meas_update}
				\begin{align}
					\bi{\chi}_{0,t \mid t} &= \bi{\chi}_{0,t \mid t-1} +  \bi{\phi}(\bi{x}_{t})\bi{\xi}_{t}\tr, \\
					\bi{\chi}_{1,t \mid t} &= \bi{\chi}_{1,t \mid t-1} +  \bi{\phi}(\bi{x}_{t}) \bi{\phi}(\bi{x}_{t})\tr, \\
					\bi{\chi}_{2,t \mid t} &= \bi{\chi}_{2,t \mid t-1} +  \bi{\xi}_{t} \bi{\xi}_{t}\tr, \\
					\chi_{3,t \mid t} &= \chi_{3,t \mid t-1} + 1.
				\end{align}
			\end{subequations}
		\end{thm}
		\begin{pf}
			See Appendix A2.
		\end{pf}
		To encode prior knowledge about the parameters, suitable initial distribution parameters can be defined $\bi{\Theta}_{0}=\{\bi{M}_{0}, \bi{V}_{0}, \bi{\Psi}_{0}, \nu_{0}\}$, as will be discussed for specific example systems in Section \ref{ch:results}.\\
		If the parameters $\bi{\theta}_t$ are known to vary over time, a common approach is to employ exponential forgetting. Doing this, historical state and interface variable estimates have less impact on the current parameter posterior. For the current model structure, exponential forgetting can be employed in the time update step of the statistics according to
		\begin{equation}\label{eq:online_time_update}
			\begin{aligned}
				\bi{\chi}_{0,t \mid t-1} &= \gamma \bi{\chi}_{0,t-1 \mid t-1}, & \bi{\chi}_{1,t \mid t-1} &= \gamma \bi{\chi}_{1,t-1 \mid t-1},\\
				\bi{\chi}_{2,t \mid t-1} &= \gamma \bi{\chi}_{2,t-1 \mid t-1}, & \chi_{3,t \mid t-1} &= \gamma \chi_{3,t-1 \mid t-1},
			\end{aligned}
		\end{equation}
		where $0 \leq \gamma \leq 1$ is the forgetting factor \cite{Berntorp.2021}.
		
		\subsection{Inference: The state and interface variables}\label{ch:online_inference}
		To estimate the state and interface variables, we approximate the density 
		\begin{equation}\label{eq:online_state_posterior}
			\begin{aligned}
				p(\bi{x}_{0:t}, \bi{\xi}_{0:t}|\bi{y}_{0:t}) \approx \sum_{i=1}^{N} q_{t}^i \delta_{\{ \bi{x}_{0:t}^i, \bi{\xi}_{0:t}^i\}}(\bi{x}_{0:t}, \bi{\xi}_{0:t}),
			\end{aligned}
		\end{equation}
		by a set of $N$ weighted particles, each representing a state and corresponding interface variable trajectory. The importance weight of particle $i$ in time step $t$, associated with trajectories $\bi{x}_{0:t}^i, \bi{\xi}_{0:t}^i$, is denoted $q_t^{i}$. The trajectories are obtained by recursive sampling from a tractable proposal distribution $\pi (\bi{x}_t, \bi{\xi}_t | \bi{x}_{0:t-1}, \bi{\xi}_{0:t-1},\bi{y}_{0:t})$. The weights $\{q_{t}^i\}_{i=1}^{N}$ are updated according to 
		\begin{equation}
			q^{i}_{t} \propto \frac{p(\bi{y}_{t}| \bi{x}^{i}_{0:t}) p(\bi{x}^{i}_{t},\bi{\xi}^{i}_{t}|\bi{x}^{i}_{0:t-1}, \bi{\xi}^{i}_{0:t-1})}{\pi(\bi{x}^{i}_{t}, \bi{\xi}^{i}_{t}| \bi{x}^{i}_{0:t-1}, \bi{\xi}^{i}_{0:t-1}, \bi{y}_{0:t})} q^{i}_{t-1}.
		\end{equation}
		Here, we exploit that we can condition the proposal distribution on $\bi{y}_t$ and use an auxiliary \ac{pf} \cite{Sarkka.2023,Wills.2023}, resulting in the weight update
		\begin{equation}\label{eq:online_weight_update_full}
			q^{i}_{t} \propto \frac{p(\bi{y}_{t}| \bi{x}^{i}_{0:t}) p(\bi{x}^{i}_{t},\bi{\xi}^{i}_{t}|\bi{x}^{a_t^{i}}_{0:t-1}, \bi{\xi}^{a_t^{i}}_{0:t-1})}{\pi_t(\bi{x}^{i}_{t}, \bi{\xi}^{i}_{t}, a_t^{i} | \bi{y}_{0:t})} q^{a_t^{i}}_{t-1},
		\end{equation}
		with the ancestor indices $a_t^{i}$ and new samples drawn from the generating proposal distribution
		\begin{subequations}\label{eq:online_proposal}
			\begin{align}
				&\pi_t (\bi{x}_t, \bi{\xi}_t, a_t | \bi{y}_{0:t}) \propto \pi_t (\bi{x}_t, \bi{\xi}_t | a_t, \bi{y}_{0:t}) \pi_t ( a_t | \bi{y}_{0:t}),\label{eq:online_proposal_a}\\
				& \nonumber\\
				&\pi_t (\bi{x}_t, \bi{\xi}_t | a_t, \bi{y}_{0:t}) = p(\bi{x}_{t},\bi{\xi}_{t}|\bi{x}^{a_t}_{0:t-1}, \bi{\xi}^{a_t}_{0:t-1}), \label{eq:online_proposal_c}\\
				&\pi_t ( a_t | \bi{y}_{0:t}) \propto   \lambda^{a_t},\label{eq:online_proposal_b}
			\end{align}
		\end{subequations}
		where resampling is done based on the auxiliary weights $\lambda^{a_t}$. In a bootstrap particle filter, $\lambda^{a_t} = q^{a_t}_{t-1}$ for all particles. In an auxiliary particle filter, the optimal importance distribution is approximated by
		\begin{equation}
			\lambda^{a_t} = q^{a_t}_{t-1} p(\bi{y}_{t}| \tilde{\bi{x}}^{a_t}_{t}),
		\end{equation}
		incorporating knowledge about $\bi{y}_t$ in the propagation step. Here, the auxiliary state $\tilde{\bi{x}}^{a_t}_{t}$ is chosen as the mean of $p(\bi{x}_{t},\bi{\xi}_{t}|\bi{x}^{a_t}_{0:t-1}, \bi{\xi}^{a_t}_{0:t-1})$. Thus, the weight update \eqref{eq:online_weight_update_full} simplifies to
		\begin{equation}\label{eq:online_weight_update}
			q^{i}_{t} \propto \frac{p(\bi{y}_{t}| \bi{x}^{i}_{0:t})}{p(\bi{y}_{t}| \tilde{\bi{x}}^{a_t^{i}}_{t})}=\frac{\mathcal{N}(\bi{y}_t | \bi{h}({\bi{x}}_t^{i}) ,\bi{\Sigma}_e)}{\mathcal{N}(\bi{y}_t | \bi{h}(\tilde{\bi{x}}_t^{a_t^{i}}) ,\bi{\Sigma}_e)}.
		\end{equation}
		So far, a standard auxiliary \ac{pf} that recursively performs weight update, resampling, and propagation has been described. As we employ a marginalized \ac{pf}, however, drawing samples from \eqref{eq:online_proposal_c} depends on the current parameter posterior and requires further attention. For marginalization of the parameters $\bi{\theta}$, we expand the likelihood  \eqref{eq:online_proposal_c} and integrate out the parameters particle-based, according to
		\begin{equation}\label{eq:online_marginalization}
			\begin{aligned}
				&p(\bi{x}_{t},\bi{\xi}_{t}|\bi{x}_{0:t-1}, \bi{\xi}_{0:t-1}) \\
				&= p(\bi{x}_{t}|\bi{x}_{t-1}, \bi{\xi}_{t-1}) p(\bi{\xi}_{t}|\bi{x}_{t},\bi{\eta}_t)\\
				&= p(\bi{x}_{t}|\bi{x}_{t-1}, \bi{\xi}_{t-1}) \int p(\bi{\xi}_{t}|\bi{x}_{t}, \bi{\theta}) p(\bi{\theta}| \bi{\eta}_t) \mathrm{d}\bi{\theta},
			\end{aligned}
		\end{equation}
		where the integrand in \eqref{eq:online_marginalization} is a hierarchical model constituting a normal and a $\mathcal{MNIW}$ distribution, 
		\begin{subequations}\label{eq:online_hierarchical_sampling}
			\begin{align}
				\bi{\xi}_{t} &\sim \mathcal{N}\left(\bi{\xi}_{t} \mid \bi{A} \bi{\phi}(\bi{x}_{t}), \bi{\Sigma}_\epsilon \right),\\
				\bi{A}, \bi{\Sigma}_{\epsilon} & \sim \mathcal{MNIW}(\bi{A}, \bi{\Sigma}_{\epsilon}| \bar{\bi{M}}, \bar{\bi{V}}, \bar{\bi{\Psi}}, \bar{\nu}), \label{eq:online_hierarchical_sampling_b}
			\end{align}
		\end{subequations}
		and the distribution parameters $\bar{\bi{M}}^{i}$, $\bar{\bi{V}}^{i}$, $\bar{\bi{\Psi}}^{i}$, $\bar{\nu}^{i}$ of the current parameter posterior of particle $i$ is determined by the statistics $\bi{\eta}_t^{i} = \{ \bi{\chi}_{0,t}^{i},\bi{\chi}_{1,t}^{i},\bi{\chi}_{2,t}^{i},{\chi}_{3,t}^{i} \}$ using \eqref{eq:MdlStr_mniw_suffstat2para}. 
		Due to the conjugate prior configuration in \eqref{eq:online_hierarchical_sampling}, the predictive density $p(\bi{\xi}_{t}|\bi{x}_{t},\bi{\eta}_t)$ is a matrixvariate Student-t ($\mathcal{MT}$) distribution. To avoid computationally expensive sampling from a matrix distribution, the density can be transformed to a multivariate Student-t ($\mathcal{T}$) distribution, following the lines of \cite{Berntorp.2021}.
		\begin{thm}\label{thm:Online_student_t}
			Given the hierarchical model \eqref{eq:online_hierarchical_sampling} and corresponding distribution parameters $\bar{\bi{M}}$, $\bar{\bi{V}}$, $\bar{\bi{\Psi}}$, $\bar{\nu}$, the predictive distribution of the interface variable $\bi{\xi}_t$ can be written as
			\begin{equation}\label{eq:online_predictive_distribution}
					p(\bi{\xi}_{t}|\bi{x}_{t},\bi{\eta}_t) = \mathcal{T}(\bi{\xi}_{t}| \bar{\rho},\bar{\bi{\mu}},\bar{\bi{\Lambda}} ),
			\end{equation}
			with distribution parameters
			\begin{subequations}\label{eq:online_predictive_distribution_parameters}
				\begin{align}
					\bar{\rho} &= \bar{\nu} - n_{\xi} +1, & \bar{\bi{\mu}} &= \bar{\bi{M}}\bi{\phi}(\bi{x}_t),\\
					\bar{\bi{\Lambda}} &= \frac{\bar{\kappa} + 1}{\bar{\kappa} \bar{\rho}} \bar{\bi{\psi}}, \label{eq:scale_matrix} & \bar{\kappa} &= 1/\left(\bi{\phi}(\bi{x}_t)\tr \bar{\bi{V}}\bi{\phi}(\bi{x}_t)\right).\\
				\end{align}
			\end{subequations} 
		\end{thm}
		\begin{pf}
			See Appendix A3. 
		\end{pf}
		\begin{rmk}
			In Theorem \ref{thm:Online_student_t}, the computation of the scale matrix $\bar{\bi{\Lambda}}$ differs from the proposition in \cite[Theorem 3]{Berntorp.2021}. Taking a different path in the proof, it is shown that the scale matrix can be constructed by \eqref{eq:scale_matrix}, resulting in a computationally more efficient implementation, as no Cholesky factorizations are required to construct $\bar{\bi{\Lambda}}$.
		\end{rmk}

		\section{Offline inference and learning}\label{ch:methods_offline}
		For offline inference and learning, we are given a measurement trajectory $\bi{y}_{0:T}$ and target the joint smoothing posterior 
		\begin{equation}\label{eq:offline_posterior}
			\begin{aligned}
				p(\bi{x}_{0:T}, \bi{\xi}_{0:T}, \bi{\theta}|\bi{y}_{0:T}) = p(\bi{\theta}|\bi{x}_{0:T}, \bi{\xi}_{0:T}) p(\bi{x}_{0:T}, \bi{\xi}_{0:T}|\bi{y}_{0:T}),
			\end{aligned}
		\end{equation}
		where the parameters $\bi{\theta}$ are modeled time-invariant. As direct sampling from the high-dimensional density $p(\bi{x}_{0:T}, \bi{\xi}_{0:T}, \bi{\theta}|\bi{y}_{0:T})$ is intractable, we use \ac{pmcmc} for data augmentation \cite{Andrieu.2010,Wigren.2022}. Specifically, marginalized \ac{pgas} \cite{Wigren.2019} is a highly efficient identification algorithm that allows to exploit the chosen model structure \eqref{eq:MdlStr_ModelEquationsSSM} advantageously.\\
		Marginalized \ac{pgas} belongs to the family of \ac{pg} methods \cite{Andrieu.2010} which sample from an intractable joint density by recursively sampling from its marginal distributions. In the current system identification setting, this corresponds to recursively drawing (i) trajectory samples $\{\bi{x}_{0:T}[k], \bi{\xi}_{0:T}[k]\}$ from the density
		\begin{equation}\label{eq:offline_state_posterior}
			\begin{aligned}
				p(\bi{x}_{0:T}, \bi{\xi}_{0:T}|\bi{y}_{0:T}),
			\end{aligned}
		\end{equation}
		using \ac{csmc} and (ii) sampling parameters $\bi{\theta}[k]$ from the density
		\begin{equation}\label{eq:offline_param_posterior}
			\begin{aligned}
				p(\bi{\theta}|\bi{x}_{0:T}, \bi{\xi}_{0:T}),
			\end{aligned}
		\end{equation}
		for iterations $k=1,\dots,K$. However, standard \ac{pg} samplers often suffer from slow exploration, meaning that large numbers of iterations $K$ are needed to sample from the target density, which is usually referred to as ``poor mixing'' of a \ac{pg} sampler \cite{Wigren.2022}.\\
		In this context, marginalized \ac{pgas} is an extension that significantly improves mixing without greatly increasing the number of required particles. The main adjustments regarding standard \ac{pg} are an ancestor sampling (AS) step \cite{Lindsten.2014} and parameter marginalization \cite{Wigren.2019} in the \ac{csmc} part. For details regarding these extensions, we refer the reader to \cite{Lindsten.2014} and \cite{Wigren.2019}, respectively.\\ 
		The resulting inference and learning scheme will by construction asymptotically provide samples from the true posterior density $p(\bi{\theta} | \bi{y}_{0:T})$ \cite{Andrieu.2010,Svensson.2017}. However, the length of the burn-in period cannot be guaranteed. To mitigate this issue, mixing can be improved by decreasing the risk of path degeneracy. Instead of using a large number of particles, we rely on and derive an auxiliary version of marginalized \ac{csmc}.\\
		The offline inference and learning scheme is summarized in Algorithm~\ref{alg:offline_mPGAS}, and the details are spelled out in the following subsections.
		\begin{algorithm}[tb]
			\caption{Offline inference and learning of \eqref{eq:MdlStr_ModelEquationsSSM}}
			\label{alg:offline_mPGAS}
			\textbf{Input:} Data $\bi{y}_{0:T}$, parameter prior $p(\bi{\theta})$.\\
			\textbf{Output:} $K$ MCMC samples of the invariant distribution \eqref{eq:offline_posterior}.
			\begin{algorithmic}[1]
				\State Set $ \bi{x}_{0:T}[0], \bi{\xi}_{0:T}[0]$ arbitrarily.
				\State Draw $\bi{\theta}[0] \: | \: \bi{x}_{0:T}[0], \bi{\xi}_{0:T}[0]$. \Comment{by \eqref{eq:offline_parameter_posterior}}
				\For{$k = 1,\dots,K$}
				\State Draw  \Comment{Algorithm~\ref{alg:offline_mcsmc}}
				\Statex \hspace{4mm} $\bi{x}_{0:T}[k], \bi{\xi}_{0:T}[k] \: | \: \bi{x}_{0:T}[k\!-\!1], \bi{\xi}_{0:T}[k\!-\!1]$. 
				\State Calculate parameter posterior \Comment{by \eqref{eq:offline_suffstats_update}, \eqref{eq:offline_suffstats}}
				\State Draw \hspace{5.8mm} $\bi{\theta}[k] \: | \: \bi{x}_{0:T}[k], \bi{\xi}_{0:T}[k]$. \Comment{by \eqref{eq:offline_parameter_posterior}} 
				\EndFor
			\end{algorithmic}
		\end{algorithm}
		
		\subsection{Learning: The parameter posterior}
		Generating parameter samples corresponds to sampling from \eqref{eq:offline_param_posterior}, which is, due to Bayes' rule,
		\begin{equation}\label{eq:offline_parameter_posterior_simple}
			p(\bi{\theta}|\bi{x}_{0:T}, \bi{\xi}_{0:T}) \propto p(\bi{x}_{0:T}, \bi{\xi}_{0:T}| \bi{\theta}) p(\bi{\theta}),
		\end{equation}
		with the likelihood
		\begin{equation}\label{eq:offline_parameter_decomposition}
			\begin{aligned}
				&p(\bi{x}_{0:T}, \bi{\xi}_{0:T}| \bi{\theta}) \\
				&\quad = p(\bi{x}_0)  \prod_{t=0}^{T-1} p(\bi{x}_{t+1} \mid  \bi{x}_{t}, \bi{\xi}_{t}) \underbrace{ \prod_{t=0}^{T}  p( \bi{\xi}_{t} \mid \bi{x}_{t}, \bi{\theta}   ) }_{p(\bi{\xi}_{0:T} | \bi{x}_{0:T}, \bi{\theta})}.
		\end{aligned}
	\end{equation}
	As we assume that the state transition function \eqref{eq:state_trans} is known and, thus, independent of $\bi{\theta}$, \eqref{eq:offline_parameter_decomposition} is proportional to $p(\bi{\xi}_{0:T} | \bi{x}_{0:T}, \bi{\theta})$ regarding the parameters. For clarity, let us further assume $p(\bi{x}_0)$ to be known and omit it. With known parameters and due to the Gaussian noise model $p( \bi{\xi}_{t} \mid \bi{x}_{t}, \bi{\theta}   ) = \mathcal{N}\left(\bi{\xi}_{t} \mid \bi{A} \bi{\phi}(\bi{x}_{t}), \bi{\Sigma}_\epsilon \right)$, the log density is, after some manipulations,
	\begin{equation}
		\begin{aligned}
			&\log p(\bi{\xi}_{0:T} | \bi{x}_{0:T}, \bi{\theta})  =	-\frac{{s}_3+n_\xi}{2} \log 2\pi -\frac{{s}_3}{2} \log |\bi{\Sigma}_\epsilon| \\
			&\quad -\frac{1}{2} \Tr\left(\bi{\Sigma}_\epsilon^{-1}\left(\bi{s}_2 -\bi{A} \bi{s}_0-\bi{s}_0\tr \bi{A}\tr+\bi{A} \bi{s}_1 \bi{A}\tr\right)\right),
		\end{aligned}
	\end{equation}
	with the ``new'' trajectory statistics $\bi{s}_0$, $\bi{s}_1$, $\bi{s}_2$, ${s}_3$ \cite{Svensson.2017}. Now, from \eqref{eq:offline_parameter_posterior_simple} and due to the choice of a conjugate prior, the parameter posterior can be calculated in closed form by summation of the prior statistics $\bi{\eta}_0$ and the ``new'' statistics $\bi{s}_0$, $\bi{s}_1$, $\bi{s}_2$, ${s}_3$, which is provided by the following Theorem.  
	\begin{thm}\label{thm:Offline}
		Given model structure \eqref{eq:MdlStr_ModelEquationsSSM}, parameter prior $p(\bi{\theta})$ with distribution parameters $\bi{\Theta}_0=\{\bi{M}_0, \bi{V}_0, \bi{\Psi}_0, \nu_0\}$, and sample trajectories $\bi{x}_{0:T}$, $\bi{\xi}_{0:T}$, the posterior distribution of the parameters is
		\begin{equation}\label{eq:offline_parameter_posterior}
			\begin{aligned}
				&p(\bi{\theta}|\bi{x}_{0:T}, \bi{\xi}_{0:T}) \\
				&\qquad =\mathcal{MNIW}(\bi{A}, \bi{\Sigma}_{\epsilon}| \bi{M}^{+}, \bi{V}^{+}, \bi{\Psi}^{+}, \nu^{+}),
			\end{aligned}
		\end{equation}
		with $\bi{M}^{+}$, $\bi{V}^{+}$, $\bi{\Psi}^{+}$, $\nu^{+}$ being calculated by \eqref{eq:MdlStr_mniw_suffstat2para}, and the prior sufficient statistics $\bi{\eta}_0 = \{ \bi{\chi}_{0,0},\bi{\chi}_{1,0},\bi{\chi}_{2,0},{\chi}_{3,0} \}$ being calculated by \eqref{eq:MdlStr_mniw_para2suffstat}, if the corresponding statistics updates are
		\begin{equation}\label{eq:offline_suffstats_update}
			\begin{aligned}
				&\bi{\chi}_{0}^+ = \bi{\chi}_{0,0} + \bi{s}_0, \qquad
				&&\bi{\chi}_{1}^+ = \bi{\chi}_{1,0} + \bi{s}_1, \\
				&\bi{\chi}_{2}^+ = \bi{\chi}_{2,0} + \bi{s}_2, 
				&&\chi_{3}^+ = \chi_{3,0} + {s}_3,
			\end{aligned}
		\end{equation}
		with the trajectory statistics
		\begin{equation}\label{eq:offline_suffstats}
			\begin{aligned}
				&\bi{s}_0 = \sum_{t=0}^{T} \bi{\phi}(\bi{x}_{t})\bi{\xi}_{t}^{\top}, \quad
				&&\bi{s}_1  =  \sum_{t=0}^{T} \bi{\phi}(\bi{x}_{t}) \bi{\phi}(\bi{x}_{t})\tr, \\
				&\bi{s}_2  = \ \sum_{t=0}^{T} \bi{\xi}_{t} \bi{\xi}_{t}^{\top}, 
				&&{s}_3  = \sum_{t=0}^{T} 1.
			\end{aligned}
		\end{equation}
	\end{thm}
	\begin{pf}
		The result extends Theorem \ref{thm:Online} for sample trajectories $\bi{x}_{0:T}$, $\bi{\xi}_{0:T}$ instead of single samples $\bi{x}_{t}$, $\bi{\xi}_{t}$. From \eqref{eq:offline_parameter_decomposition}, it is clear that the likelihood $p(\bi{\xi}_{0:T} | \bi{x}_{0:T}, \bi{\theta})$ is a product of Gaussian distributions, resulting in a summation of the statistics of the considered exponential family member.
	\end{pf}

	\subsection{Inference: The state and interface variables}
		To draw the state and interface variables in Algorithm\,\ref{alg:offline_mPGAS}, we use \ac{smc} and approximate the density 
		\begin{equation}\label{eq:offline_state_posterior}
			\begin{aligned}
				p(\bi{x}_{0:T}, \bi{\xi}_{0:T}|\bi{y}_{0:T}) \approx \sum_{i=1}^{N} q_{T}^i \delta_{\{ \bi{x}_{0:T}^i, \bi{\xi}_{0:T}^i\}}(\bi{x}_{0:T}, \bi{\xi}_{0:T}),
			\end{aligned}
		\end{equation}
		by a set of $N$ weighted particles, each representing a state and corresponding interface variable trajectory. The resulting marginalized \ac{pf} is similar to the one described in Section \ref{ch:online_inference} with an additional conditioning step to ensure sampling from the correct target distribution. The resulting scheme is summarized in Algorithm \ref{alg:offline_mcsmc}.\\
		In more detail, the trajectories are drawn using \ac{csmc}, which resembles a standard \ac{pf} with one particle being fixed to given reference trajectories $\bi{x}_{0:T}'$, $\bi{\xi}_{0:T}'$ \cite{Wigren.2022}. 
		In the current setting, we draw the ancestor $a_t^{N}$ of the deterministically set samples $\bi{x}_t^{N} = \bi{x}_t'$ and $\bi{\xi}_t^{N} = \bi{\xi}_t'$ by considering the joint probability of the concatenated trajectories $\left[\bi{x}_{0: t}^i, \bi{x}'_{t+1: T}\right]$ and $\left[\bi{\xi}_{0: t}^i, \bi{\xi}'_{t+1: T}\right]$ 
		\begin{equation}\label{eq:offline_ancestor_sampling_general}
			\begin{aligned}
				\tilde{q}_{t \mid T}^i ={\lambda}^i \frac{p(\left[\bi{x}_{0: t}^i, \bi{x}'_{t+1: T}\right],\left[\bi{\xi}_{0: t}^i, \bi{\xi}'_{t+1: T}\right], \bi{y}_{0: T})}{p(\bi{x}_{0: t}^i, \bi{\xi}_{0: t}^i, \bi{y}_{0: t})},
			\end{aligned}
		\end{equation}
		for each particle $i$. Following the lines of \cite{Wigren.2019}, the ancestor weights in marginalized \ac{pgas} are
		\begin{equation}\label{eq:offline_ancestor_sampling}
			\begin{aligned}
				\tilde{q}_{t \mid T}^i \propto {\lambda}^i \frac{g(\bi{\eta}_{t}^i)}{g(\bi{\eta}_T^i)} h_{t+1}^i,
			\end{aligned}
		\end{equation}
		where $h_{t+1}$ is the data-dependent base measure of the likelihood, and $g(\cdot)$ is the normalizing constant of the parameter posterior, dependent on the current statistics $\bi{\eta}_t$ or $\bi{\eta}_T$, respectively. The terms $g(\cdot)$ and $h_{t+1}$ are derived by expressing the respective distributions in the canonical form of the restricted exponential family, and, for completeness, the derivation is outlined in Appendix A1.
		\begin{algorithm}[tb]
			\caption{\\Marginalized conditional \ac{smc} (for all $i=1,\dots,N$)}
			\label{alg:offline_mcsmc}
			\textbf{Input:} Data $\bi{y}_{0:T}$, reference trajectories $ \bi{x}_{0:T}', \bi{\xi}_{0:T}'$, reference statistics $\bi{s}_0',\bi{s}_1',\bi{s}_2',{s}_3'$, parameter prior $p(\bi{\theta})$.\\
			\textbf{Output:} Sample new trajectories $\bi{x}_{0:T}[k]$, $\bi{\xi}_{0:T}[k]$.
			\begin{algorithmic}
				\State Draw $\bi{x}_{0}^{1:N-1}, \bi{\xi}_{0}^{1:N-1} \sim p(\bi{x}_{0}, \bi{\xi}_{0})$
				\State Set $\bi{x}_{0}^N = \bi{x}_{0}'$ and $\bi{\xi}_{0}^N = \bi{\xi}_{0}'$.
				\State Set weights $q^{1:N}_{0} = 1/N$.
				\For{$t = 1,\dots,T$}
				\State Update statistics $\bi{s}_{1:4,t+1:T}'$. \Comment{by \eqref{eq:offline_marg_stat_ref_t}}		
				\State Update hyperparameters $\bi{\eta}_T^{i}$, $\bi{\eta}_t^{i}$. \Comment{by \eqref{eq:offline_marg_stat_T}, \eqref{eq:offline_marg_stat_t}}
				\State Generate auxiliary states $\tilde{\bi{x}}_t^{i} = \bi{f}(\bi{x}_{t-1}^{i},\bi{\xi}_{t-1}^{i})$.
				\State Compute first-stage weights 
				\Statex \hspace{4mm} $\lambda^{i} \propto q_{t-1}^{i} \mathcal{N}(\bi{y}_t | \bi{h}(\tilde{\bi{x}}_t^{i}) ,\bi{\Sigma}_e)$ and normalize.
				\State Draw $a_t^{i} \sim \mathcal{C}(\{\lambda^{i}\}_{i=1}^{N})$.
				\State Draw $a_t^{N} \sim \mathcal{C}(\{\tilde{q}_{t-1|T}^{i}\}_{i=1}^{N})$. \Comment{by \eqref{eq:offline_ancestor_sampling}}
				\State Draw $\bi{x}_t^{i} \sim \mathcal{N}(\bi{x}_t|\bi{f}(\bi{x}_{t-1}^{a_t^{i}},\bi{\xi}_{t-1}^{a_t^{i}}), \bi{\Sigma}_\omega)$.
				\State Set $\bi{x}_t^{N} = \bi{x}_t'$.
				\State Draw $\bi{\xi}_t^{i} \sim p(\bi{\xi}_t | \bi{x}_t^{i}, \bi{\eta}_t^{a_t^{i}})$. \Comment{by \eqref{eq:online_predictive_distribution}}
				\State Set $\bi{\xi}_t^{N} = \bi{\xi}_t'$.
				\State Compute weights ${q}_t^{i}$ and normalize. \Comment{by \eqref{eq:online_weight_update}}
				\EndFor
				\State Sample $J$ with $\mathbb{P}(J = j) \propto {q}_T^{i}$
				\State Set $\bi{x}_{0:T}[k]=\bi{x}_{0:T}^{J}$, $\bi{\xi}_{0:T}[k]= \bi{\xi}_{0:T}^{J}$.
			\end{algorithmic}
		\end{algorithm}
		Specifically, the concatenated sufficient statistics $\bi{\eta}_T^{i}= \{ \bi{\chi}_{0,T}^{i},\bi{\chi}_{1,T}^{i},\bi{\chi}_{2,T}^{i},{\chi}_{3,T}^{i} \}$ 
		are computed by
		\begin{subequations}\label{eq:offline_marg_stat_T}
			\begin{align}
				&\bi{\chi}_{0,T}^{i} = \bi{\chi}_{0,t}^{i} + \bi{s}_{0,t+1:T}', &\bi{\chi}_{1,T}^{i} = \bi{\chi}_{1,t}^{i} + \bi{s}_{1,t+1:T}', \\ 
				&\bi{\chi}_{2,T}^{i} = \bi{\chi}_{2,t}^{i} + \bi{s}_{2,t+1:T}', &\chi_{3,T}^{i} = \chi_{3,t}^{i} + {s}_{3,t+1:T}',
			\end{align}
		\end{subequations}
		where the current statistics $\bi{\eta}_t^{i}= \{ \bi{\chi}_{0,t}^{i},\bi{\chi}_{1,t}^{i},\bi{\chi}_{2,t}^{i},{\chi}_{3,t}^{i} \}$ for particle $i$ in time step $t$
		\begin{subequations}\label{eq:offline_marg_stat_t}
			\begin{align}
				&\bi{\chi}_{0,t}^{i} = \bi{\chi}_{0,t-1}^{i} + \bi{\phi}(\bi{x}_{t}^{i}) \bi{\xi}_{t}^{i\top}, \\
				&\bi{\chi}_{1,t}^{i} = \bi{\chi}_{1,t-1}^{i} + \bi{\phi}(\bi{x}_{t}^{i}) \bi{\phi}(\bi{x}_{t}^{i})\tr, \\
				&\bi{\chi}_{2,t}^{i} = \bi{\chi}_{2,t-1}^{i} + \bi{\xi}_{t}^{i} \bi{\xi}_{t}^{i\top}, \\
				&\chi_{3,t}^{i} = \chi_{3,t-1}^{i} + 1,
			\end{align}
		\end{subequations}
		and the remaining reference statistics
		\begin{subequations}\label{eq:offline_marg_stat_ref_t}
			\begin{align}
				&\bi{s}_{0,t+1:T}' = \bi{s}_{0,t:T}' -\bi{\phi}(\bi{x}_{t}') \bi{\xi}_{t}'^{\top}, \\
				&\bi{s}_{1,t+1:T}'  =  \bi{s}_{1,t:T}' -\bi{\phi}(\bi{x}_{t}') \bi{\phi}(\bi{x}_{t}')\tr, \\
				&\bi{s}_{2,t+1:T}'  =  \bi{s}_{2,t:T}' - \bi{\xi}_{t}' \bi{\xi}_{t}'^{\top}, \\
				&{s}_{3,t+1:T}'  =  {s}_{3,t:T}' -1.
			\end{align}
		\end{subequations}

		%
		
		\section{Experiments}\label{ch:results}
		To show the broad applicability of the proposed learning framework, we apply the algorithms in three different case studies\footnote{The corresponding code for simulations, inference and learning, as well as result visualization is available online at \href{https://github.com/VolkmannB/bayesian-inference-with-explicit-and-implicit-prior-knowledge}{\texttt{https://github.com/VolkmannB/bayesian-inference}} \href{https://github.com/VolkmannB/bayesian-inference-with-explicit-and-implicit-prior-knowledge}{\texttt{-with-explicit-and-implicit-prior-knowledge}}.}, including a real-world benchmark data set of an \ac{emps} \cite{Janot.2019}. The section is organized as follows. First, a convenient choice for a basis function expansion \eqref{eq:basis_func_exp} is introduced to incorporate implicit prior knowledge about the target functions to be learned. Second, the testbeds are introduced briefly. Last, the respective results for offline and online inference and learning are presented and discussed.
		\subsection{Choosing a basis function expansion} \label{ch:results_basisFunctions}
		For initialization of the proposed learning framework, it is necessary to define suitable basis functions $\bi{\phi}(\cdot)$ and a $\mathcal{MNIW}$ prior for the parameters $\bi{\theta} = \{\bi{A}, \bi{\Sigma}_\epsilon\}$. Doing this, implicit prior knowledge about the target function can be incorporated to facilitate learning \cite{Berntorp.2022,Svensson.2017}.\\
		From expert knowledge about the considered system, basis functions that span a desired input domain, satisfy certain symmetry properties, or that encode known physical relationships can be defined \cite{Champion.2020}. Moreover, data-driven approaches to find suitable basis functions are available, \eg using \cite{Brunton.2016}. The $\mathcal{MNIW}$ prior can be used to set a desired weighting for the respective basis functions.\\
		A particularly elegant way to incorporate implicit prior knowledge into learning while admitting for uncertainty is to choose and parameterize a suitable kernel in a \ac{gp} model \cite{Rasmussen.2005}. This can be cast into the functional form \eqref{eq:basis_func_exp} using a parametric \ac{gp} approximation presented in \cite{Solin.2020,RiutortMayol.2023}, which is based on a Hilbert-space representation. For Bayesian inference and learning of \ac{gp} state-space models, this model structure has been exploited previously in \cite{Berntorp.2021,Svensson.2017}.\\
		In the case studies, we make use of the reduced-rank \ac{gp} proposed in \cite{Solin.2020}. To retain conciseness, we introduce only the resulting basis function approximation here, further details regarding the GP approximation can be found in \cite{Solin.2020,RiutortMayol.2023}. 
		%
		Using \cite{Solin.2020}, implicit prior knowledge about the target function is embedded in the current interface variable model \eqref{eq:basis_func_exp} by constructing a set of basis functions
		\begin{align}
			\bi{\phi}(\bi{x}_t) &= \begin{bmatrix}
				\phi_k(\bi{x}_t) & \hdots
			\end{bmatrix}\tr, k = 1,\ldots,n_\phi \text{ },\\
			\phi_{k}(\bi{x}_t) &\triangleq \prod_{i=1}^{n_x} \frac{1}{\sqrt{L_i}} \sin \left(\frac{\pi j_i\left(x_{t,i}+L_i\right)}{2 L_i}\right),
		\end{align}
		defined on a domain $\bi{x}_t \in \Omega_x = [-L_1,L_1] \times \ldots \times [-L_{n_x}, L_{n_x}] \in \mathbb{R}^{D}$, where a combination of integers $(j_1, \hdots, j_{n_x})$ is chosen to maximize the expressiveness of each basis function. The column covariance $\bi{V}$ of the function prior in \eqref{eq:MdlStr_MNIW_Prior} is defined as
		\begin{align}
			\bi{V} = \mathrm{diag}\left(S\left(\sqrt{\varrho_1}\right),\hdots,S\left(\sqrt{\varrho_{n_\phi}}\right)\right),	
		\end{align}
		where $S(\cdot)$ is the spectral density function of the chosen kernel, and $\varrho_1, \hdots, \varrho_{n_\phi}$ are the eigenvalues of the basis functions.

		\subsection{Testbeds and evaluation criteria}
		For the experiments, we employ i) a single-mass oscillator simulation in which an unknown nonlinear spring and damper characteristic is learned, ii) a lateral vehicle dynamics simulation in which the tire-friction characteristic is learned, and iii) a real-world \ac{emps} \cite{Janot.2019} in which velocity-dependent resistance forces are learned. In all examples, inference and learning are accomplished based on using input-output data only. For completeness, the respective system models and experiment setups are illustrated in the Appendix (see page \pageref{ap:spring_damper}).\\
		As the first two case studies are based on simulations, the learning success is quantified using the \ac{rmse} between the function posteriors mean $\bi{M}\bi{\phi}(\bi{x})$ and the ground truth $\bi{\Xi}(\bi{x})$ over the function input domain $\Omega_x$. Considering this domain, the learned approximation $\bi{A}\bi{\phi}(\bi{x})$ is less accurate in poorly explored regions (\ie few training data points), and the associated variance is relatively high. For a fair error quantification over the whole function domain, while taking into account limited exploration during the experiment, we employ the \ac{wrmse} 
		\begin{equation}\label{eq:wRMSE3}
				e_{i,\mathrm{wRMSE}} = \sqrt{ c \int_{\Omega_x} w_i(\bi{x}) [\bi{\Xi}(\bi{x}) - \bi{M}\bi{\phi}(\bi{x})]_{i}^2 \mathrm{d}\bi{x} },
		\end{equation}
		for the error of the $i$-th learned function in multi-task regression, where the weighting $w_i(\bi{x})$ is the inverse marginal variance of the posterior \cite{Murphy.2007} at the respective input $\bi{x}$, \ie
		\begin{equation}\label{eq:weighting}
			w_i(\bi{x}) = \frac{\nu}{\bi{\phi}(\bi{x})\tr {\bi{V}} \bi{\phi}(\bi{x}) [\bi{\Psi}]_{ii}},
		\end{equation}
		and $c$ is a normalizing constant to keep track of the correct units. For evaluation, the integral in \eqref{eq:wRMSE3} is numerically approximated using equidistant points over the entire input domain $\Omega_x$.\\
		In the case of the experimental \ac{emps} benchmark, the ground truth of the target function is unknown. Therefore, we resort to evaluating the \ac{rmse} between the state estimates and the available measurement data, allowing for a direct performance comparison with \cite{Champneys.2024,Forgione.2021}. 
		
		\begin{figure*}[h]
			\centering
			\includegraphics[width=1\textwidth]{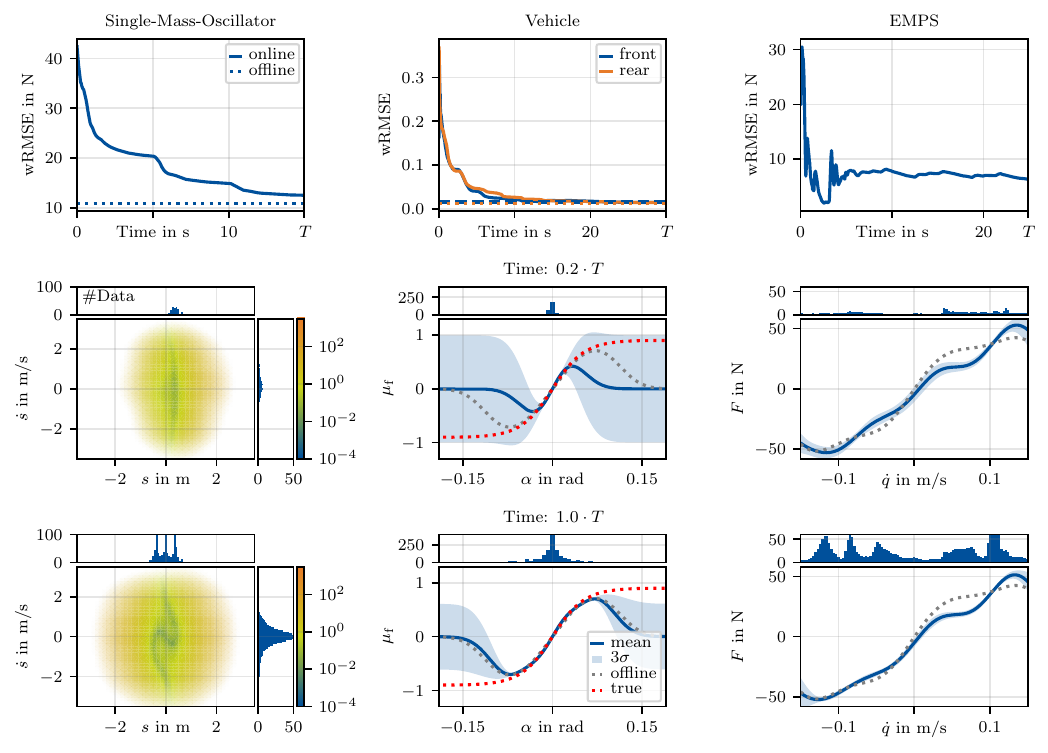}
			\caption{Results for the learned functions from Algorithm\,\ref{alg:online_RBPF}. 
				The first row shows the \ac{wrmse} to the known target function. 
				In the second and third rows, the learned target functions at time $t=0.2\cdot T$ and $t=1.0\cdot T$ are shown, respectively. 
				For the single-mass oscillator, a heat map of the absolute mean error with logarithmic color scale is shown. 
				In all examples, a decrease of the \ac{wrmse} over time is visible. The function error decreases proportional to the number of data points in the given input region. This highlights, that the proposed framework can be applied to a wide set of scenarios, where unknown models may have uni- or multivariate inputs or even multiple models in parallel.
			}
			\label{fig:online_results_fcn}
		\end{figure*}
		
		\subsection{Initialization}
		In all examples, zero-mean Gaussian process and measurement noise are assumed. 
		All \ac{smc} methods employ $200$ particles, and all marginalized \ac{pgas} runs use $800$ iterations. In general, the reference trajectories in Algorithm~\ref{alg:offline_mPGAS} can be initialized arbitrarily. As an convenience to avoid long burn-in periods, the first reference trajectory for marginalized \ac{pgas} is initialized with a trajectory from the filtering distribution, obtained by running Algorithm \ref{alg:online_RBPF}.\\
		For initialization of the remaining distribution parameters, noninformative priors $\bi{M} = \bi{0}$, $\bi{\Psi} = a\bi{\mathrm{I}}_{n_\xi}$ and $\nu = n_\xi$ are set. The covariance kernel hyperparameters are tuned manually, and the respective values can be found in the appendix.\\
		All case studies utilize explicit prior knowledge in the form of state dynamics equations derived from first principles and implicit prior knowledge about the target function by utilizing basis functions according to section\,\ref{ch:results_basisFunctions}.
		In the vehicle simulation, we target learning of the tire-road friction characteristic. From expert knowledge, we know that this relationship is typically anti-symmetric with respect to the origin \cite{Pacejka.2012}. This knowledge has been exploited in \cite{Berntorp.2022} to constrain the employed \ac{gp} approximation and thus, to reduce the number of required basis functions. In the present case study, we follow a similar approach and exploit this expert knowledge by setting only anti-symmetric basis functions in the vehicle simulation. Additionally the basis functions take not the system state as input, but rather the vehicles side-slip angle, derived from the system state.

		\begin{figure*}[h]
			\centering
			\includegraphics[width=1\textwidth]{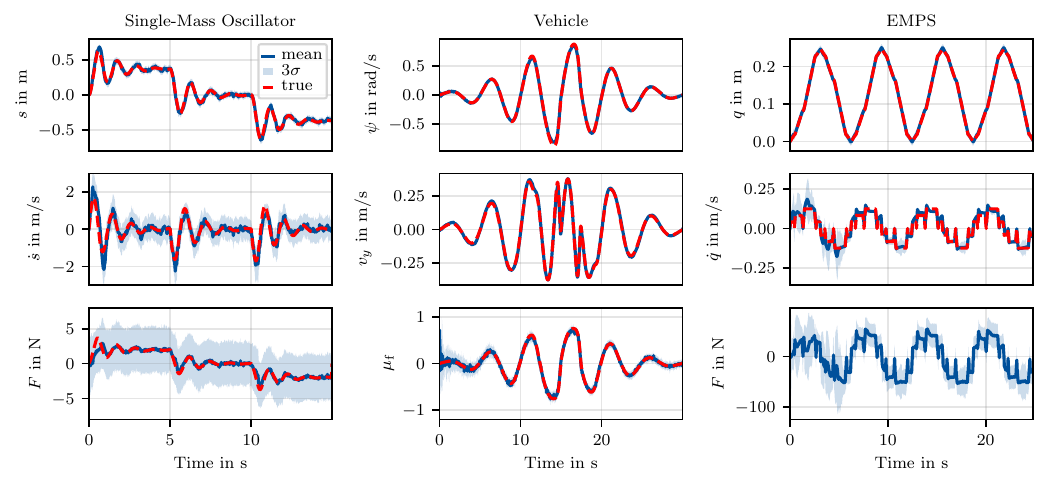}
			\caption{
				State trajectories $\bi{x}_{0:T}$ of the single-mass oscillator, the vehicle simulation, and the \ac{emps} (top and middle row) generated from Algorithm\,\ref{alg:online_RBPF}. The bottom plots depict the trajectories of the interface variables $\xi_{0:T}$. 
				In all examples, a learning effect in the estimation accuracy of states and interface variables is visible. Moreover, the uncertainty of the estimates decreases over the considered time interval.
			}
			\label{fig:online_results_trj}
		\end{figure*}
		\begin{figure*}[h]
			\centering
			\includegraphics[width=1\textwidth]{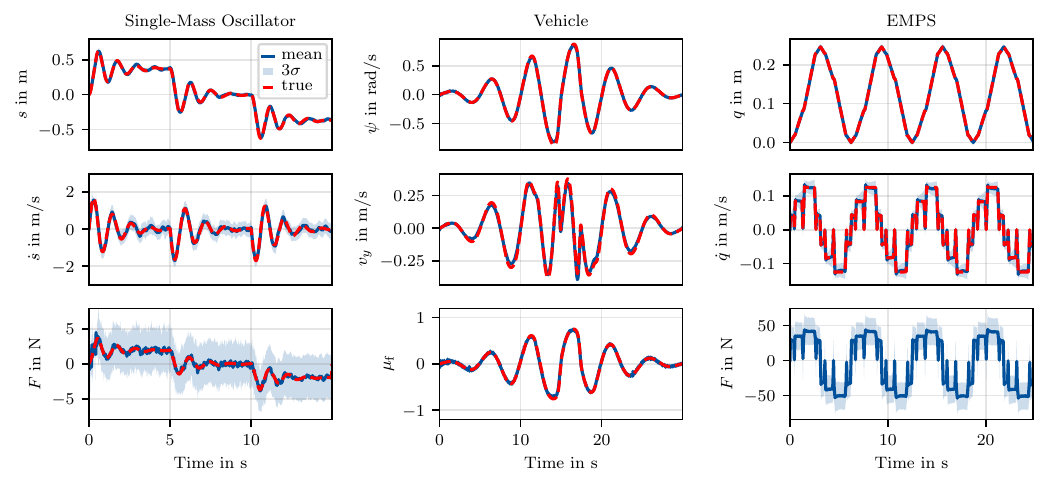}
			\caption{
				State trajectories $\bi{x}_{0:T}$ of the single-mass oscillator, the vehicle simulation, and the \ac{emps} (top and middle row) generated from Algorithm\,\ref{alg:offline_mPGAS}. The bottom plots depict the trajectories of the interface variables $\xi_{0:T}$. 
				The results are generally similar to the online case. Compared to Fig.\,\ref{fig:online_results_trj} the trajectories of $\bi{x}_{0:T}$ and $\xi_{0:T}$ show less noise since the samples were drawn from the smoothing distribution.
			}
			\label{fig:offline_results_trj}
		\end{figure*}
		\subsection{Online inference and learning results}\label{sec:results_online}
		Fig.\,\ref{fig:online_results_fcn} shows the learned target function and the associated uncertainty interval for different time steps, when running Algorithm \ref{alg:online_RBPF}. The histograms next to the target function plots indicate the number of data points $\{\bi{x}_t,\bi{\xi}_t\}$ in the corresponding input domain region, used for learning. It is worth mentioning that the data points are not measured directly but are inferred by marginalized particle filtering in Algorithm~\ref{alg:online_RBPF}. Please note, the single-mass oscillator system (left column) exhibits an input domain in two dimensions. Thus, the \ac{rmse} with respect to the true target function is depicted as a heat map.\\
		As can be seen, the online learning scheme converges to the target function shapes as more data arrives over time. The approximation of the target function is usually good in input domain regions with a sufficient number of data points. On the other hand, the accuracy of the learned function degrades, and its uncertainty increases in unexplored regions. Here, no information regarding the target function is contained in the data, and the \ac{gp} approximation stays at its (noninformative) prior, which can be seen clearly in the vehicle simulation results (middle column).\\
		The learning effect is quantified using the \ac{wrmse} \eqref{eq:wRMSE3} in the top row of Fig.\,\ref{fig:online_results_fcn}. All target function errors using Algorithm~\ref{alg:online_RBPF} decrease rapidly. For comparison, the final error measures of the computationally significantly more demanding offline Algorithm \ref{alg:offline_mPGAS} are indicated. As can be seen, Algorithm~\ref{alg:online_RBPF} comes close to the learning accuracy of offline \ac{pmcmc} but does not reach it which can be attributed to the forgetting factor employed in Algorithm~\ref{alg:online_RBPF}.\\
		As the true target function in the \ac{emps} benchmark is not known, we judge the learning accuracy by quantifying the \ac{wrmse} relative to an offline learned target function (see Section \ref{sec:results_offline}). Similar to the simulation case studies, the error decreases with time, which suggests a learning capability for the real-world \ac{emps} benchmark \cite{Janot.2019} as well.\\
		Moreover, the learning effect is visible in the trajectory estimates, depicted in Fig.\,\ref{fig:online_results_trj}. The top and middle plots show the state estimates, while the bottom plots show the interface variable estimates. As can be seen for all three case studies, the estimation accuracy of states and interface variables improves over the considered time horizon. In addition, the uncertainty intervals decrease, as more data becomes available and the input domain is explored.

		\begin{figure*}[h]
			\centering
			\includegraphics[width=1\textwidth]{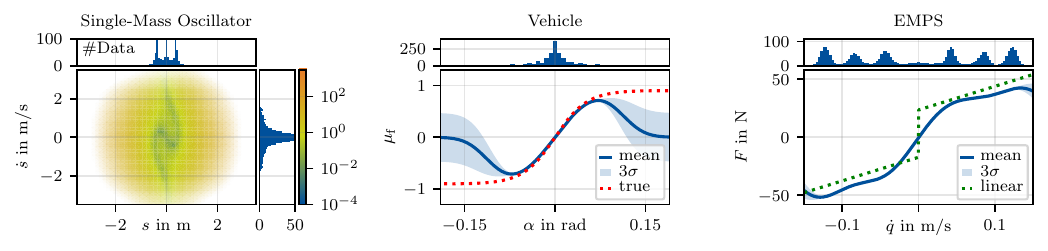}
			\caption{
				Results for the learned functions from Algorithm\,\ref{alg:offline_mPGAS}. 
				For the single-mass oscillator, a heat map shows the mean absolute error at a logarithmic scale. 
				For the vehicle, the red line shows the ground truth of the friction curve. For the \ac{emps}, the red line shows a linear resistance model using viscous and Coulomb friction as well as an offset, obtained by least squares estimation. 
				In contrast to the offline case, Algorithm\,\ref{alg:offline_mPGAS} assumes constant model parameters $\boldsymbol{\theta}$ and applies no exponential forgetting. As a result, the remaining error is slightly lower when compared to the results of Algorithm\,\ref{alg:offline_mPGAS} in Fig.\,\ref{fig:online_results_fcn}.
			}
			\label{fig:offline_results_fcn}
		\end{figure*}
		\subsection{Offline inference and learning results}\label{sec:results_offline}
		In Fig.\,\ref{fig:offline_results_trj}, the smoothed state and interface variables, resulting from Algorithm~\ref{alg:offline_mPGAS} are depicted for all three case studies. As expected, the estimation accuracy is improved, and the uncertainty intervals are smaller compared to the results of the online inference and learning scheme. Moreover, the accuracy and uncertainty intervals are comparable along the \textit{time} axis, as learning is accomplished over \textit{iterations}.\\
		The learned function approximations after $800$ iterations are shown in Fig.\,\ref{fig:offline_results_fcn}. As can be seen, the target function in the simulation examples is learned accurately in all regions of available data. Please note, high accuracy is achieved even in regimes with few data points, as all data is taken into account when learning over \textit{iterations} without forgetting.\\
		Regarding the learnable function shapes, Fig.\,\ref{fig:offline_results_fcn} suggests that even highly nonlinear functions can be approximated accurately, using the presented learning framework. In particular, a forward simulation with the learned target function using the test trajectory in \cite{Janot.2019} revealed a position \ac{rmse} of \SI{8.4}{mm}. For comparison, a simulation with a piecewise linear friction model considering only Coulomb and viscous damping yields \SI{9.0}{mm} \ac{rmse}.
		%

		\section{Conclusion and outlook}\label{ch:conclusion}
		In recent years, diverse and powerful methods for nonlinear system identification have been developed, but the fusion of measurement data information with diverse prior knowledge, \eg explicitly known system equations and implicit assumptions about the function to be learned, is subject to current research. In particular, a pressing research question is how different types of prior knowledge can be fused to cope with limited data.\\
		As a response, we propose a novel general-purpose identification tool for inference of latent states and learning of unknown system dynamics from input-output data, while incorporating explicit and implicit prior knowledge. Leveraging forward sampling by \ac{smc} methods, no user-defined coordinate transformations or model inversions are needed.  Specifically, the tool relies on an interchangeable \textit{interface} between a priori known system equations and a learning-based representation for unknown model parts. Data-efficient learning is enabled by incorporating implicit prior knowledge about the target function in a \ac{gp} approximation.\\
		For efficient offline and online inference and learning, closed form expressions for the parameter distribution are derived and exploited in marginalized \ac{smc} methods. 
		The effectiveness and broad applicability of the proposed learning framework are shown in three distinct case studies, including an experimental electro-mechanical positioning system benchmark \cite{Janot.2019}. 
		Thus, the contributions represent a step towards flexible and data-efficient inference and learning in complex real-world systems by enabling the use of diverse 
		prior knowledge about the system.\\
		%
		%
		Considering future research, the algorithms might be employed for inference and learning in relevant real-world applications, \eg for adaptive control systems in road and rail vehicles, wind compensation in aerial vehicles, or friction- and wear-adaptive industrial machines. 
		Moreover, a potentially fruitful research direction is to incorporate more general implicit knowledge, \eg energy conservation laws, in the learning-based representation as proposed in \cite{Cranmer.2019,Beckers.2022}.

		
		\bibliographystyle{plain}        
		\bibliography{references}           
		
		\section*{Appendix}

		\subsection*{A1. The canonical form of restricted exponential family} \label{ap:canonical_form}    
		Writing a specific distribution in the canonical form of the exponential family allows to employ general results that apply to all members of the exponential family. In particular, we consider the restricted exponential family which is convenient for use in state-space models, as dependence on the previous state can be incorporated \cite{Wigren.2019}.\\
		In the current setting, we strive to express the likelihood as
		\begin{align}\label{eq:appendix_lik}
			& p(\bi{x}_t, \bi{\xi}_t \mid \bi{x}_{t-1}, \bi{\xi}_{t-1}, \bi{\alpha} )\\
			&\quad = p(\bi{\xi}_t, \mid \bi{x}_{t}, \bi{\alpha} ) p(\bi{x}_t\mid \bi{x}_{t-1}, \bi{\xi}_{t-1} )\\ 
			&\quad = \mathcal{N}(\bi{\xi}_{t}|\bi{A}\bi{\phi}(\bi{x}_t), \bi{\Sigma}_{\epsilon})\mathcal{N}(\bi{x}_{t}|\bi{f}(\bi{x}_{t-1},\bi{\xi}_{t-1}), \bi{\Sigma}_{\omega})\\
			& \quad = {h}_t \exp \left(\bi{\alpha}^{\top} \bi{s}(\bi{x}_t,\bi{\xi}_t) - \bi{P}^{\top}(\bi{\alpha}) \bi{r}(\bi{x}_t)\right), \nonumber
		\end{align}
		and the conjugate prior as
		\begin{align}\label{eq:appendix_conj_prior}
			p\left(\bi{\alpha} \mid \bi{s}_0, \bi{r}_0\right)&= \mathcal{MNIW}(\bi{A},\bi{\Sigma}_{\epsilon}|\bi{M}, \bi{V}, \bi{\Psi}, \nu)\\
			&={g} \left(\bi{s}_0, \bi{r}_0\right) \exp \left(\bi{\alpha}^{\top} \bi{s}_0-\bi{P}^{\top}(\bi{\alpha}) \bi{r}_0\right), \nonumber
		\end{align}
		with the natural parameters $\bi{\alpha}$ that are linked to the parameters of interest $\bi{\theta}$ by some transformation. Defining
		\begin{equation}
			\bi{\alpha} =  \begin{bmatrix}
				\bi{\alpha}_0 \\ \bi{\alpha}_1
			\end{bmatrix} = \begin{bmatrix}
				\bi{\Sigma}_{\epsilon}^{-1} \bi{A} \\ -\frac{1}{2} \bi{\Sigma}_{\epsilon}^{-1}
			\end{bmatrix},
		\end{equation}
		and rearranging terms in \eqref{eq:appendix_lik} and \eqref{eq:appendix_conj_prior} yields the parameter-dependent part of the log-partition function
		\begin{align}
			\bi{P}(\bi{\alpha}) &= \begin{bmatrix}
				-\frac{1}{4} \bi{\alpha}_0\tr \bi{\alpha}_1^{-1}\bi{\alpha}_0 \\ -\frac{1}{2} \log |-2\bi{\alpha}_1 |
			\end{bmatrix} = \begin{bmatrix}
				\frac{1}{2} \bi{A}\tr\bi{\Sigma}_{\epsilon}^{-1}\bi{A} \\ \frac{1}{2} \log |\bi{\Sigma}_{\epsilon} |
			\end{bmatrix}.
		\end{align}
		For the likelihood \eqref{eq:appendix_lik}, the statistics are
		\begin{align}
			\bi{s}(\bi{x}_t,\bi{\xi}_t) &= \begin{bmatrix}
				\bi{\phi}(\bi{x}_t)\bi{\xi}_t\tr \\ \bi{\xi}_t\bi{\xi}_t\tr
			\end{bmatrix}, \\
			\bi{r}(\bi{x}_t) &= \begin{bmatrix}
				\bi{\phi}(\bi{x}_t)\bi{\phi}(\bi{x}_t)\tr \\ 1
			\end{bmatrix} ,
		\end{align}
		and the base measure 
		\begin{equation}
			{h}_t =  (2\pi)^{-n_\xi/2} \mathcal{N}(\bi{x}_{t}|\bi{f}(\bi{x}_{t-1},\bi{\xi}_{t-1}), \bi{\Sigma}_{\omega}).
		\end{equation}
		For the $\mathcal{MNIW}$ distribution \eqref{eq:appendix_conj_prior}, the sufficient statistics are
		\begin{align}
			\bi{s}_0 &= \begin{bmatrix}
				\bi{\chi}_{0} \\ \bi{\chi}_2
			\end{bmatrix} = \begin{bmatrix}
				\bi{V}^{-1} \bi{M} \\  \bi{M}\bi{V}^{-1}\bi{M}^{-1} + \bi{\Psi}
			\end{bmatrix}, \\
			\bi{r}_0 &= \begin{bmatrix}
				\bi{\chi}_{1} \\ {\chi}_3
			\end{bmatrix} = \begin{bmatrix}
				\bi{V}^{-1} \\ \nu 
			\end{bmatrix},
		\end{align}
		which reflects the relationship between sufficient statistics and distribution parameters in Lemma \ref{def:tranformation_suff_stats}, and for defining the initial statistics, $\nu_0 = n_{\xi} + n_{\phi} + 1$ should be set.\\
		Please note, computing the posterior now amounts to summing $\bi{s}_{\mathrm{new}} = \bi{s}_0 + \bi{s}(\bi{x}_t,\bi{\xi}_t)$ and $\bi{r}_{\mathrm{new}} = \bi{r}_0 + \bi{r}(\bi{x}_t)$, which resembles the update rules in Theorem \ref{thm:Online}.\\
		The normalizing factor $\bi{g}(\bi{s}_0, \bi{r}_0)$, relevant for ancestor sampling in \eqref{eq:offline_ancestor_sampling}, is
		\begin{equation}
			{g}(\bi{s}_0, \bi{r}_0) = \frac{|\bi{\Psi}|^{\nu/2}}{2^{\nu n_{\xi}/2} (2\pi)^{n_\xi n_{\phi}/2} |\bi{V}|^{n_\xi /2}  \bi{\Gamma}_{n_{\xi}}(\nu / 2)}.
		\end{equation}

		\subsection*{A2. Proof sketch of Theorem \ref{thm:Online}}    
		\label{ch:App_Posterior}
		For clarity, we omit the time index $t$ in the following and reiterate that the interface variable is defined
		\begin{equation}
			\bi{\xi} = \bi{A} \bi{\phi}(\bi{x}) + \bi{\epsilon}, \qquad \bi{\epsilon} \sim \mathcal{N}(\bi{\epsilon}| \bi{0}, \bi{\Sigma}_\epsilon),
		\end{equation}
		resulting in the likelihood
		\begin{equation}\label{eq:appendix_likelihood}
			\begin{aligned}
				p( \bi{\xi} | \bi{x}, \bi{\theta}) = \mathcal{N}\left(\bi{\xi} | \bi{A} \bi{\phi}(\bi{x}), \bi{\Sigma}_\epsilon \right).
			\end{aligned}
		\end{equation}
		Using \eqref{eq:appendix_likelihood} and the parameter prior 
		\begin{equation}
			p( \bi{\theta}) = \mathcal{MNIW}(\bi{A}, \bi{\Sigma}_{\epsilon}| \bi{M}, \bi{V}, \bi{\Psi}, \nu),
		\end{equation}
		the current parameter posterior follows from Bayes' rule as
		\begin{equation}\label{eq:appendix_posterior}
			\begin{aligned}
				p(\bi{\theta} | \bi{\xi}, \bi{x}) &\propto p(\bi{\xi}| \bi{x}, \bi{\theta} ) p( \bi{\theta})\\
				& \propto \mathcal{N}(\bi{\xi}|\bi{A}\bi{\phi}(\bi{x}),\bi{\Sigma}_{\epsilon}) \mathcal{MNIW}(\bi{A}, \bi{\Sigma}_{\epsilon}).
			\end{aligned}
		\end{equation}
		Due to this conjugate prior configuration, the posterior is again a $\mathcal{MNIW}$ distribution, and rearranging the product on the right-hand side of \eqref{eq:appendix_posterior} accordingly yields update laws for the sufficient statistics \cite{Berntorp.2021,Svensson.2017}. In particular, the logarithmic posterior is
		\begin{align}
				&\log p(\bi{A}, \bi{\Sigma}_{\epsilon}| \bi{\xi}, \bi{x}) \propto \nonumber\\
				&-\frac{1}{2} \log \left| \bi{\Sigma}_{\epsilon}\right| -\frac{1}{2} \mathrm{Tr}\left((\bi{\xi} - \bi{A}\bi{\phi}(\bi{x})) (\bi{\xi} - \bi{A}\bi{\phi}(\bi{x}))\tr \bi{\Sigma}_{\epsilon}^{-1}\right) \nonumber\\
				&-\frac{1}{2}\log\left|\bi{\Sigma}_{\epsilon}\right|(n_{\phi} +n_{\xi}+ \nu +1) \\
				&-\frac{1}{2} \mathrm{Tr}\left((\bi{A}-\bi{M}) \bi{V}^{-1} (\bi{A}-\bi{M})\tr \bi{\Sigma}_{\epsilon}^{-1}\right) \nonumber\\
				&-\frac{1}{2} \mathrm{Tr}\left(\bi{\Psi} \bi{\Sigma}_{\epsilon}^{-1}\right) , \nonumber
		\end{align}
		where the first two terms on the right-hand side result from the likelihood, and the last three terms result from the prior. Rearranging and collecting terms yields
		%
		%
		\begin{equation}
			\begin{aligned}
				&\log p(\bi{A}, \bi{\Sigma}_{\epsilon}| \bi{\xi}, \bi{x}) \propto \\
				&-\frac{1}{2} \mathrm{Tr}(( - (
				\underbrace{ \bi{M}\bi{V}^{-1} + \bi{\xi} \bi{\phi}(\bi{x})\tr }_{(\bi{\chi}_{0} + \bi{\phi}(\bi{x}) \bi{\xi}\tr)\tr}
				) \bi{A}\tr ) \bi{\Sigma}_{\epsilon}^{-1}) \\
				&-\frac{1}{2}\mathrm{Tr}( ( - \bi{A} (
				\underbrace{\bi{V}^{-1}\bi{M}\tr + \bi{\phi}(\bi{x}) \bi{\xi}\tr}_{\bi{\chi}_{0} + \bi{\phi}(\bi{x}) \bi{\xi}\tr}
				)) \bi{\Sigma}_{\epsilon}^{-1}) \\
				&-\frac{1}{2}\mathrm{Tr}( ( \bi{A}(
				\underbrace{\bi{V}^{-1} + \bi{\phi}(\bi{x})\bi{\phi}(\bi{x})\tr}_{\bi{\chi}_{1} + \bi{\phi}(\bi{x})\bi{\phi}(\bi{x})\tr}
				)\bi{A}\tr ) \bi{\Sigma}_{\epsilon}^{-1})\\
				&-\frac{1}{2} \mathrm{Tr}((
				\underbrace{\bi{M}\bi{V}^{-1}\bi{M}\tr + \bi{\Psi} + \bi{\xi} \bi{\xi}\tr }_{\bi{\chi}_{2} + \bi{\xi} \bi{\xi}\tr}
				)\bi{\Sigma}_{\epsilon}^{-1})\\
				&-\frac{1}{2}\log\left|\bi{\Sigma}_{\epsilon}\right|(
				\underbrace{\nu+1}_{\chi_{3} + 1}
				+n_{\phi}+n_{\xi}+1),
			\end{aligned}
		\end{equation}
		with the indicated updates for the sufficient statistics $\bi{\eta} = \{ \bi{\chi}_{0},\bi{\chi}_{1},\bi{\chi}_{2},{\chi}_{3} \}$, considering the transformation \eqref{eq:MdlStr_mniw_para2suffstat}.
		
		\subsection*{A3. Proof sketch of Theorem \ref{thm:Online_student_t}}    
		To prove that the predictive distribution $p(\bi{\xi}_{t}|\bi{x}_{t},\bi{\eta}_t)$ follows a Student-t density with parameters \eqref{eq:online_predictive_distribution_parameters}, we note that the predictive distribution of a random variable $\bi{z} \in \mathbb{R}^{p}$ with likelihood $\mathcal{N}\left(\bi{z} \mid {\bi{\mu}}, \bi{\Sigma} \right)$ and $\mathcal{NIW}$ prior
		\begin{subequations}
			\begin{align}
				{\bi{\mu}}  & \sim \mathcal{N}({\bi{\mu}} | {\bi{\mu}}_0, \bi{\Sigma} / \kappa_0), \label{eq:appendix_multivariate_normal_example}\\
				\bi{\Sigma} & \sim \mathcal{IW}(\bi{\Sigma}| {\bi{\Lambda}}_0, {\nu}_0), 
			\end{align}
		\end{subequations}
		is the Student-t density
		\begin{equation}
			\mathcal{T}(\bi{z} | {\rho},{\bi{\mu}},{\bi{\Lambda}} ),
		\end{equation}
		with
		\begin{equation}
			\begin{aligned}
				{\rho} &= {\nu}_0 -p +1,\\
				{\bi{\mu}} &= {\bi{\mu}}_0,\\
				{\bi{\Lambda}} &= \frac{\kappa_0 + 1}{\kappa_0 \rho} {\bi{\Lambda}}_0,
			\end{aligned}
		\end{equation}
		which is a standard result (see, \eg \cite{Murphy.2007}). Now, this fact can be exploited by showing how the hierarchical model \eqref{eq:online_hierarchical_sampling} with distribution parameters $\bar{\bi{\Theta}} = \{\bar{\bi{M}}, \bar{\bi{V}}, \bar{\bi{\Psi}}, \bar{\nu}\}$ can be transformed to a $\mathcal{NIW}$ density. This idea has been employed in \cite{Berntorp.2021} as well, but taking a slightly different approach that requires the computation of four Cholesky factors. Instead, we rearrange the linear transformation by $\bi{\phi}(\bi{x}_t)$ and write the hierarchical model \eqref{eq:online_hierarchical_sampling} as
		\begin{subequations}\label{eq:appendix_hierarchical_sampling_rewritten}
			\begin{align}
				\bi{\xi}_{t} &\sim \mathcal{N}\left(\bi{\xi}_{t} \mid \bar{\bi{\xi}}, \bi{\Sigma}_\epsilon \right),\\
				\bar{\bi{\xi}}  & \sim \mathcal{MN}(\bar{\bi{\xi}}| \bar{\bi{M}} \bi{\phi}(\bi{x}_{t}), \bi{\Sigma}_{\epsilon}, \bi{\phi}(\bi{x}_{t})\tr \bar{\bi{V}} \bi{\phi}(\bi{x}_{t})), \label{eq:appendix_mn_xi}\\
				\bi{\Sigma}_{\epsilon} & \sim \mathcal{IW}(\bi{\Sigma}_{\epsilon}| \bar{\bi{\Psi}}, \bar{\nu}), 
			\end{align}
		\end{subequations}
		where the $\mathcal{MN}$ distribution in \eqref{eq:appendix_mn_xi} is a matrix distribution over a column vector $\bar{\bi{\xi}}$. Equivalently, $\bar{\bi{\xi}}$ can be sampled from a multivariate Gaussian
		\begin{equation}\label{eq:appendix_multivariate_normal}
			\begin{aligned}
				\bar{\bi{\xi}}  &\sim \mathcal{N}(\bar{\bi{\xi}} | \bar{\bi{M}} \bi{\phi}(\bi{x}_{t}), \bi{\Sigma}_{\epsilon} / \bar{\kappa} ),\\
				\bar{\kappa} &= 1/(\bi{\phi}(\bi{x}_{t})\tr \bar{\bi{V}} \bi{\phi}(\bi{x}_{t})).
			\end{aligned}
		\end{equation}
		Comparing and identifying terms in \eqref{eq:appendix_multivariate_normal_example} and \eqref{eq:appendix_multivariate_normal} concludes the proof.

		\subsection*{A4. Single-mass oscillator simulation} \label{ap:spring_damper}    
		The first example is the well-known single-mass oscillator system, where a single point mass $m$ is suspended by a spring and a damper. In the present simulation, the spring and damper characteristics are nonlinear and unknown. The states are the displacement $s$ and the velocity $\dot{s}$ of the mass, \ie $\bi{x}\tr = \left[ s,  \dot{s} \right]$. The objective is to learn the spring and damper relationship while inferring the state variables from input-output data only. The differential equations of these states are
		\begin{equation}
			\frac{\mathrm{d}}{\mathrm{d} t}\begin{bmatrix}
				s \\ \dot{s}
			\end{bmatrix} = \begin{bmatrix}
				\dot{s} \\
				\left(F_{\mathrm{ext}} - F_{\mathrm{sd}}(s,\dot{s}) \right)/m
			\end{bmatrix},
		\end{equation}
		where $u = F_{\mathrm{ext}}$ is an external input force, and $F_{\mathrm{sd}}$ is the overall spring and damper force acting on the system. The spring and damper characteristic in this numerical case study is given by
		\begin{equation}\label{eq:target_function_oscillator}
			\begin{aligned}
				F_{\mathrm{s}}(s) &= c_{1} s + c_{2} s^3 , \\
				F_{\mathrm{d}}(\dot{s}) &= d_{1} \dot{s} \frac{1}{1+d_{2} \dot{s} \tanh(\dot{s})} ,\\
				F_{\mathrm{sd}}(s, \dot{s}) &= F_{\mathrm{s}}(s) + F_{\mathrm{d}}(\dot{s}),
			\end{aligned}
		\end{equation}
		with parameters $c_{1}$ and $c_{2}$ for the spring and $d_{1}$ and $d_{2}$ for the damper.\\
		The system is discretized using 4-th order Runge-Kutta integration with sample time $\Delta t$ as \SI{0.02}{s}, and the simulation is evaluated for \SI{15}{s}. Starting at the origin, the system is excited multiple times by step force inputs. The sole measurement output of the system is the mass displacement $y=s$.\\
		For learning of the nonlinear characteristics \textit{inside} the state-space model, we approximate $F_{\mathrm{sd}} \approx \xi$. 
		%
		%
		For the \ac{gp} hyperparameters, the domain is set as $L_1 = L_2 = 7.5$ to be just slightly larger than the simulated trajectory of the point mass. The remaining parameters were chosen heuristically. 
		For the length scale $l$, a good initial guess was found by dividing the domain boundary length by the number of basis functions $n_\phi=41$. 
		The variance was set to $\sigma^2=10$ to allow a quick enough adaptation of the model to the data. The $\mathcal{IW}$ distribution was scaled with $a=40$ to facilitate sufficient exploration of the function space during the filters' burn-in period.

		\subsection*{A5. Nonlinear lateral vehicle dynamics simulation} \label{ap:vehicle_simulation}    
		The lateral vehicle dynamics simulation is based on the single-track model 
		\begin{equation}
			\begin{aligned}
				\dot{v}_y &= \frac{1}{m}\left( F_{z,\mathrm{f}} \mu_{y,\mathrm{f}} \cos\delta + F_{z,\mathrm{r}}\mu_{y,\mathrm{r}}  + F_{z,\mathrm{r}} \mu_x \sin\delta \right) - v_x \dot{\psi} ,\\
				\ddot{\psi} &= \frac{1}{I_{zz}} \left( l_\mathrm{f} F_{z,\mathrm{f}} \mu_{y,\mathrm{f}} \cos\delta - l_\mathrm{r} F_{z,\mathrm{r}} \mu_{y,\mathrm{r}} + l_\mathrm{f} F_{z,\mathrm{f}} \mu_x \sin\delta \right),
			\end{aligned}
		\end{equation}
		with the lateral velocity $v_y$ and yaw rate $\dot{\psi}$ as system states, \ie $\bi{x}\tr = [v_y, \dot{\psi} ]$. The inputs $\bi{u}$ to the system comprise the longitudinal velocity $v_x$ and the steering angle $\delta$. The quantities $F_z, l_\mathrm{f}, l_\mathrm{r}, m, I_{zz}, \mu_x$ are fixed parameters, and the subscripts $\square_{\mathrm{f}}$ and $\square_{\mathrm{r}}$ denote quantities related to the front and rear axle, respectively.\\
		The lateral friction $\bi{\mu}_{y} = [\mu_{y,\mathrm{f}}, \mu_{y,\mathrm{r}}]$ is determined by the nonlinear magic formula tire model \cite{Pacejka.2012} with
		\begin{equation}
			\begin{aligned}
				\mu_{y,i}(\bi{\alpha}) &= \bar{\mu} \sin(C \arctan(\zeta_i - E (\zeta_i - \arctan(\zeta_i)))), \\
				\zeta_i &= B\tan{\alpha_i},
			\end{aligned}
		\end{equation}
		for $i \in \{\mathrm{f}, \mathrm{r}\}$, where $\bar{\mu}$, $B$, $C$, and $E$ are the maximum, stiffness, shape, and curvature factor, respectively. The friction depends nonlinearly on the tire slide-slip angle $\alpha_i$, which in turn is dependent on the states $\bi{x}$.\\
		The system is discretized using 4-th order Runge-Kutta integration with sample time $\Delta t$ as \SI{0.02}{s}, and the simulation is evaluated for \SI{30}{s}. The measurement vector of the system $\bi{y}\tr = [\dot{v}_y, \dot{\psi}]$.\\
		In the numerical case study, the goal is to learn the unknown friction characteristics $\bi{\mu}(\bi{\alpha}) = [\mu_{y,\mathrm{f}}({\alpha}_{\mathrm{f}}), \mu_{y,\mathrm{r}}({\alpha}_{\mathrm{r}}) ]$ of the front and rear tire from input-output data. To this end, we use the side-slip angle as features for the basis function expansion and approximate $\bi{\mu} \approx \bi{\xi}$. 
		The \ac{gp} hyperparameters were mostly chosen similar to \cite{Berntorp.2021}, with 20 instead of 10 anti-symmetric basis functions.

		\subsection*{A6. Experimental benchmark: Electro-mechanical positioning system} \label{ap:positioning_system}    
		For real-world evaluation of the proposed learning framework, an open-source benchmark data set of an electro-mechanical positioning system is employed \cite{Janot.2019}. 
		The positioned cart is actuated by a DC motor with an incremental encoder, and the position $s$ is controlled using a PD controller. The state vector $\bi{x}\tr = [s, \dot{s}]$, and the model of the system dynamics
		\begin{equation}
			\frac{\mathrm{d}}{\mathrm{d} t}\begin{bmatrix}
				s \\ \dot{s}
			\end{bmatrix} = \begin{bmatrix}
				\dot{s} \\
				(\tau - F(\dot{s}))/m
			\end{bmatrix},
		\end{equation}
		where $u = \tau$ is the applied actuator force, $m$ is the cart's mass, and $F(\dot{s})$ describes the unknown resistance forces, including, \eg viscous damping or Coulomb friction. The sole output measurement $y = s$.\\
		The objective in this case study is to learn the (possibly nonlinear) resistance forces $F(\dot{s})$ from experimental input-output data while simultaneously inferring the systems' states $\bi{x}$. Thus, we focus on learning the resistance forces by approximating $F(\dot{s}) \approx \xi$ and define the remaining parameters, \ie the mass or process noise, manually. However, the proposed learning framework can be extended to infer those parameters as well. 
		The \ac{gp} hyperparameters were chosen heuristically, similar to the single-mass oscillator example. The domain size is set as $L_1 = 0.2$ to be large enough to include the joint's velocity trajectory. The length scale is set by dividing the domain boundary length by the number of basis functions $n_\phi=10$. 
		Variance $\sigma^2=20$ and scale $a=4$ are set such that the \ac{gp} can cover and explore possible ranges of the resistance force $F(\dot{s})$. Magnitudes are based on the identification results of a linear joint model with viscous and Coulomb friction.

\end{document}